\documentclass[aps,prl,twocolumn,groupedaddress]{revtex4}
\usepackage{amsmath,amssymb,amsthm,mathrsfs,amsfonts,dsfont} 
\usepackage{graphicx}
\usepackage{subfigure}
\usepackage{amsbsy}
\usepackage{bm}

\def\t{\operatorname{T}}
\def\bx{\mathbf{x}}
\def\id{\mathds{1}}
\def\a{\alpha}

\begin{document}

\title{Emergence of hierarchical modes from deep learning}
\author{Chan Li$^{1}$}
\author{Haiping Huang$^{1,2}$}
\email{huanghp7@mail.sysu.edu.cn}
\affiliation{$^{1}$PMI Lab, School of Physics,
Sun Yat-sen University, Guangzhou 510275, People's Republic of China}
\affiliation{$^{2}$Guangdong Provincial Key Laboratory of Magnetoelectric Physics and Devices,
Sun Yat-sen University, Guangzhou 510275, People's Republic of China}
\date{\today}

\begin{abstract}
	Large-scale deep neural networks consume expensive training costs, but the training results in less-interpretable weight matrices constructing the networks.
	Here, we propose a mode decomposition learning that can interpret the weight matrices as a hierarchy of latent modes. These modes are akin to patterns
	in physics studies of memory networks, but the least number of modes increases only logarithmically with the network width, and becomes even a constant when the width further grows.
	The mode decomposition learning not only saves a significant large amount of training costs, but also explains the network performance with
	the leading modes, displaying a striking piecewise power-law behavior.  
	The modes specify a progressively compact latent space across the network hierarchy, making a more disentangled subspaces compared to standard training.
	Our mode decomposition learning is also studied in an analytic on-line learning setting, which reveals multi-stage of learning dynamics with a continuous specialization of hidden nodes.
	Therefore, the proposed mode decomposition learning 
	points to a cheap and interpretable route towards the magical deep learning. 
\end{abstract}

 \maketitle

\textit{Introduction.}---
Deep neural networks are dominant tools with a broad range of applications in not only image and language processing,
but also scientific researches~\cite{DL-2016,RMP-2019}. These networks are parameterized by a huge amount of trainable weight matrices, thereby consuming 
an expensive training cost. However, these weight matrices are hard to interpret, and thus mechanisms underlying the macroscopic performance of the networks remain
a big mystery in theoretical studies of neural networks~\cite{HH-2022,DLT-2022}.

To save the computational cost, previous studies of deep networks applied singular value decomposition to the weight matrices~\cite{Jade-2014,SVD-2020,NC-2021,PRE-2021}.
This decomposition requires the orthogonality condition for the singular vectors and positive singular values. The training also involves a carefully-designed
structure for the trainable decomposition scheme~\cite{NC-2021,PRE-2021}. These constraints and designs make the training process complicated, and thus a concise physics interpretation is still lacking.
In addition,
previous studies of recurrent memory networks showed that
the network weight can be decomposed into separate random orthogonal patterns with corresponding importance scores~\cite{Jiang-2021a,Jiang-2021b}. Inspired by these studies,
we conjecture that the learning in deep networks is shaped by a hierarchy of latent modes, which are not necessarily orthogonal, 
and the weight matrix can be expressed by these modes.

The mode decomposition learning (MDL)
leads to a progressively compact latent mode space across the network hierarchy,
and meanwhile the subspaces corresponding to different types of input are strongly disentangled, facilitating discrimination.
The least number of latent modes achieving the comparable performance with the costly standard methods grows
only logarithmically with the network width and even could be a constant, thereby reducing significantly the training cost.
The mode spectrum exhibits an intriguing
piecewise power-law behavior. In particular, these properties do not depend on details of the training setting.
Therefore, our proposed MDL calls for a rethinking of conventional weight-based
deep learning through the lens of cheap and interpretable mode-based learning.

\textit{Model.}---
To show the effectiveness of the MDL scheme, we train a deep network to implement a classification task
of handwritten digits~\cite{mnist}. The deep network has $L$ layers ($L-2$ hidden layers) with $N_l$ neurons in the $l$-th layer.
The weight value of the connection from the neuron $i$ at the upstream layer $l$ to the neuron $j$ at the downstream layer $l+1$ is specified
by $w_{ij}^l$. The activation of the neuron $j$ at the downstream layer $h_j^{l+1}=f(z_j^{l+1})=\max(0,z_j^{l+1})$, where
the pre-activation $z_j^{l+1} = \sum_i w_{ij}^{l}h_{i}^{l}$. For the output layer, the softmax
function $h_{k}=e^{z_{k}} / \sum_{i} e^{z_{i}}$ is chosen to specify
the probability over all classes of the input images. The cross entropy $\mathcal{C}=-\sum_{i} \hat{h}_{i} \ln h_{i}$ is
used as the cost function for the supervised learning, and $\hat{h}_i$ is the target label (one-hot form). 
After training (the cross entropy is repeatedly averaged over mini-batches of training examples), 
we evaluate the generalization performance of the network on an
unseen test dataset.

Single weight values are not interpretable. According to our hypothesis, latent patterns would emerge from training in each layer.
We call these patterns hierarchical modes for deep learning.
Therefore, the relationship between the modes and weight values is expressed by the following mode decomposition,
\begin{equation}\label{mdl}
	\mathbf{w}^l = \boldsymbol{\hat{\xi}}^l\boldsymbol{\Sigma}^l (\boldsymbol{{\xi}}^{l+1})^{\t},
\end{equation} 
where there are $p^l$ upstream modes $\boldsymbol{\hat{\xi}}^l \in \mathbb{R}^{ N_l \times p^l}$, and the same number of downstream modes $\boldsymbol{\xi}^{l+1}\in \mathbb{R}^{ N_{l+1} \times p^l}$.
The importance of each pair of adjacent modes is specified by the diagonal of the importance matrix $\boldsymbol{\Sigma}^l\in\mathbb{R}^{p^l\times p^l}$.
These modes may not be orthogonal with each other, and the importance score can take a real value. This setting allows for more 
degrees of freedom for learning features of input-output mappings.
We will detail their geometric and physical interpretations below.

\begin{figure}
\centering
     \includegraphics[bb=2 3 565 421, width=0.5\textwidth]{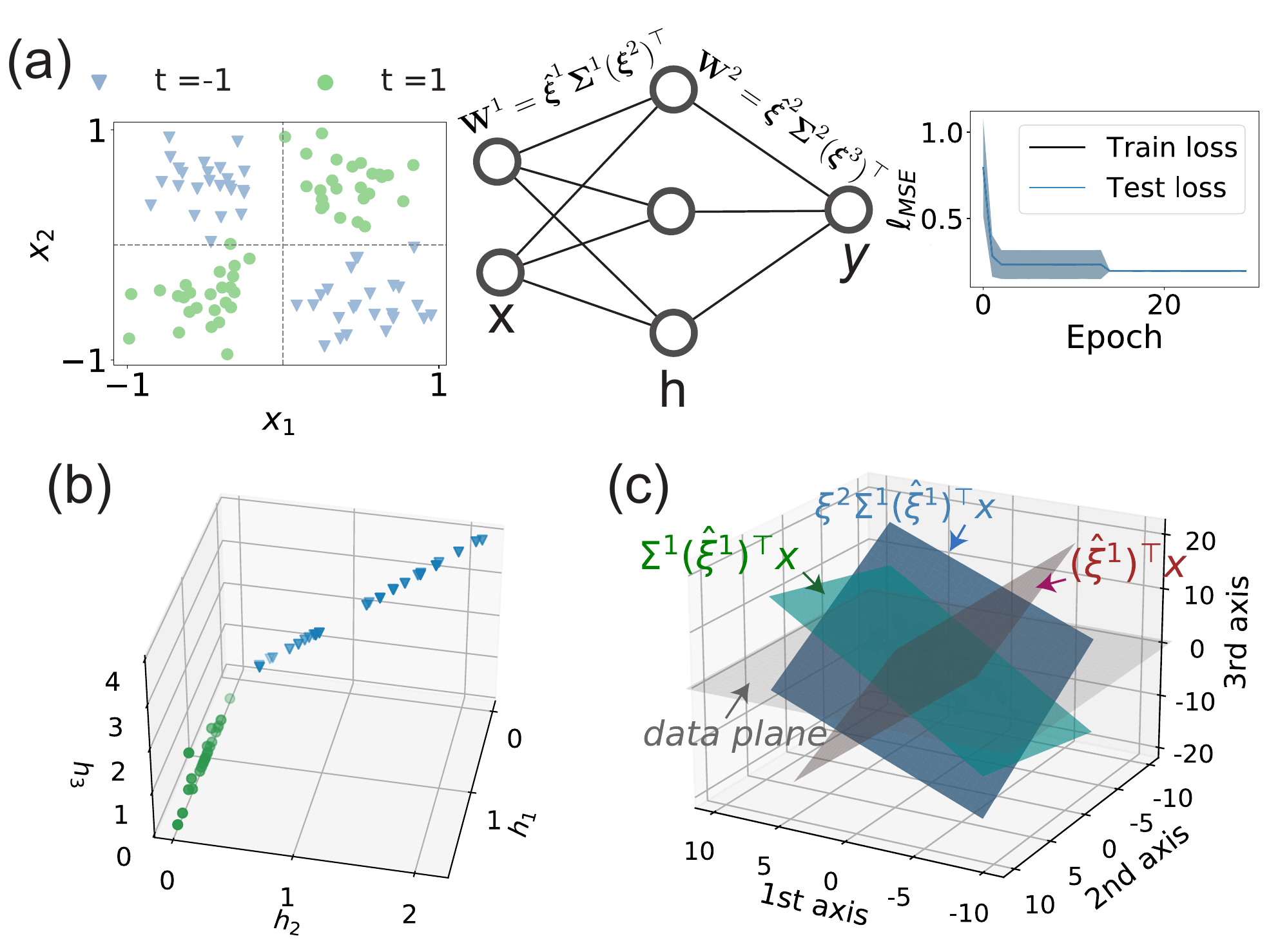}
  \caption{
 A simple illustration of the mode decomposition learning. 
 (a) A deep neural network of three layers, including one hidden layer with three hidden nodes, for a classification task of non-linearly separable data.
 The weight matrix $w_{ij}^l = \sum_{\alpha=1}^p\hat{\xi}_{i,\alpha}^l\Sigma_{\alpha}^l{\xi}_{j,\alpha}^l$, where $p=3$. 
 The distribution of input data is modeled as a Gaussian mixture (see the main text) from which 
 samples are assigned to labels $t = \pm 1$ based on the corresponding mixture component. 
  The training performance is measured by the mean-squared-error loss function $\ell_{\operatorname{MSE}}(y, t)=\|y-t\|^{2}/2$.
 (b) The representation of hidden neurons $\mathbf{h}$ plotted in the 3D space, displaying the geometric separation. (c) The successive mappings from 
 input sample $\bx$ (grey) to $(\boldsymbol{\hat{\xi}}^1)^{\t} \bx$ (dark red), followed by
 $\boldsymbol{\Sigma}^1(\boldsymbol{\hat{\xi}}^1)^{\t} \bx$ (green), and finally
 $\boldsymbol{\xi}^2\boldsymbol{\Sigma}^1(\boldsymbol{\hat{\xi}}^1)^{\t}\bx$ (blue).
  }\label{fig1}
\end{figure}

A geometric interpretation of Eq.~(\ref{mdl}) in a simple learning task is shown in Fig.~\ref{fig1}. We use a three-layer network with three hidden neurons. The input data is sampled from
a four-component Gaussian mixture~\cite{Helias-2022},
\begin{equation}
\mathbb{P}(\bx, t)=P(t) \sum_{\pm} P_{\pm} \mathcal{N}\left(\bx|\mu_{x}^{t, \pm}, \Sigma_{x}^{t, \pm}\right),
\end{equation}
where $ \mathcal{N}\left(\bx|\mu_{x}^{t, \pm}, \Sigma_{x}^{t, \pm}\right)$ denotes a Gaussian distribution with mean $\mu_{x}^{t, \pm}$ 
and covariances $\Sigma_{x}^{t, \pm}$, and $P(t) = P_{\pm} = \frac{1}{2}$. 
For the label $t=+1$, $\mu_{x}^{t=+1, \pm}=\pm(0.5,0.5)^{\t}$, while for $t=-1$,
 $\mu_{x}^{t=-1, \pm}=\pm(-0.5,0.5)^{\t}$. Covariances are isotropic throughout with $\Sigma_{x}^{t, \pm}=0.05 \id$.
The input samples $\bx\in\mathbb{R}^{2}$ are
first projected to the input pattern space spanned by $(\hat{\boldsymbol{\xi}}^1)^{\t}_i$ ($i = 1, 2,3$). 
Then all three directions of this projection get expanded or contracted via
$\boldsymbol{\Sigma}^1(\boldsymbol{\hat{\xi}}^1)^{\t}\bx$. Finally the geometrically modified representation is re-mapped to the downstream representation space of a higher dimensionality, as
$\boldsymbol{{\xi}}^2\boldsymbol{\Sigma}^1(\boldsymbol{\hat{\xi}}^1)^{\t}\bx$ [Fig.~\ref{fig1} (c)].  The non-linearity of the transfer function is then applied to the last linear transformation, leading to 
the geometric separation [Fig.~\ref{fig1} (b)]. We conclude that the MDL provides rich angles to look at the geometric transformation of the input information along the hierarchy of deep networks.

 Rather than the conventional weight values in standard backpropagation (BP) algorithms~\cite{DL-2016}, the trainable
parameters are latent patterns in the MDL.
The training is implemented by stochastic gradient descent in the mode space
$\boldsymbol{\theta}^l = (\boldsymbol{\hat{\xi}}^l, \boldsymbol{\Sigma}^l, \boldsymbol{{\xi}}^{l+1})$~\cite{SM},
\begin{equation}\label{mdleq}
\begin{aligned}
	\Delta \xi^{l+1}_{j \a}& \equiv -\eta\frac{\partial \mathcal{L}}{\partial \xi^{l+1}_{j \a}} =- \eta \mathcal{K}_j^{l+1}\Sigma_{\a}^l\sum_{i}\hat{\xi}_{i\a}^{l}h_i^{l},\\
	\Delta \Sigma_{\a}^l& \equiv -\eta\frac{\partial \mathcal{L}}{\partial \Sigma^l_{\a}} = -\eta\sum_j \mathcal{K}_j^{l+1}\xi_{j\a}^{l+1}\sum_i  \hat{\xi}_{i \a }^l h_{i}^l,\\
	\Delta \hat{\xi}^l_{i \a } & \equiv-\eta \frac{\partial \mathcal{L}}{\partial\hat{\xi}_{i \a}^{l}} =- \eta \Sigma_{\a}^lh_i^l\sum_j \mathcal{K}_j^{l+1} \xi_{j \a}^{l+1}, \\
\end{aligned}
\end{equation}
where $\mathcal{L}$ denotes the cost function (e.g., cross-entropy or mean-squared error) over a mini-batch of training data,
$\eta$ denotes the learning rate, and
$\mathcal{K}_{j}^{l+1} \equiv\partial \mathcal{L} / \partial z_{j}^{l+1}$ denotes the error term, which could back-propagate from the top layer where 
$\mathcal{K}_j^L = -\hat{h}_{j}^{L}\left(1-h_{j}^{L}\right)$ for $\mathcal{L}=\mathcal{C}$ (cross entropy). Based on the chain rule, the error backpropagation equation can be derived as 
$\mathcal{K}_i^l=\sum_j\mathcal{K}_j^{l+1}\sum_\a\xi_{i\a}^{l+1}\Sigma_\a^{l}\hat{\xi}_{j\a}^{l}f'(z_i^l)$~\cite{SM}.
To ensure the pre-activation is independent of the upstream-layer width, we take the initialization scheme that
$[\boldsymbol{{\xi}}^{l+1}\boldsymbol{{\Sigma}}^l(\hat{\boldsymbol{\xi}}^l)^{\t}]_{ij}\sim \mathcal{O}(\frac{1}{\sqrt{N_l}})$~\cite{Jiang-2021a}. 
To avoid the ambiguity of choosing patterns (e.g., scaled by a factor), we impose an identical regularization with strength $10^{-4}$
for all trainable parameters. However, our result does not change qualitatively with the specific values of regularization~\cite{SM}.   

We remark that for each hidden layer, there exist two types of pattern ($\boldsymbol{\xi}^l\neq\hat{\boldsymbol{\xi}}^l$). Equation~\eqref{mdleq} is used to learn these patterns.
We call this case 1L2P. If we assume $\boldsymbol{\xi}^l=\hat{\boldsymbol{\xi}}^l$, the training can be further simplified as in~\cite{SM}, and we call this case 1L1P.
The nature of this mode-based-computation can be understood as an expanded linear-nonlinear layered computation, as
$f(z_j^{l+1})=f(\sum_{\alpha}c_{\alpha j}\kappa_\alpha)$ where the linear field
$\kappa_\alpha=\sum_i\hat{\xi}_{i\alpha}^lh_i^l$ and the equivalent weight $c_{\alpha j}=\xi_{j\alpha}^{l+1}\Sigma_\alpha^l$. Therefore, the number of 
modes acts as the linear-layer width. We leave a systematic exploration of this linear-nonlinear structure by statistical mechanics in forthcoming works.

\textit{On-line learning dynamics in a shallow network.}---
The MDL can be analytically understood in an on-line learning setting, where we consider one-hidden-layer architecture. The on-line learning can be considered as a special case of 
the above mini-batch learning (i.e., the batch size is set to one, and the sample is visited by the learning only once).
 The training dataset consists of $n$ pairs $\left\{\bx^{\nu}, y^{\nu}\right\}_{\nu=1}^{n}$. Each training example is independently
 sampled from a probability distribution $\mathbb{P}(\bx,y)=\mathbb{P}(y|\bx)\mathbb{P}(\bx)$, where
 $\mathbb{P}(\bx)$ is a standard Gaussian distribution, and the scalar label ${y}^{\nu}$ is
 generated by the neural network of $k$ hidden neurons, (i.e., teacher, indicated by the symbol $*$ below). 
 Given an input $\bx^{\nu}\in \mathbb{R}^d$, the corresponding label is created by
 \begin{equation}
 	y^{\nu}=\frac{1}{k} \sum_{r=1}^{k} \sigma\left(\frac{[\boldsymbol{{\xi}}^*\boldsymbol{{\Sigma}^{*}}(\boldsymbol{\hat{\xi}^{*}})^{\t}]_{r} \bx^{\nu}}{\sqrt{d}}\right)=\frac{1}{k} \sum_{r=1}^{k} \sigma\left(\lambda_{r}^{* \nu}\right), 
 \end{equation}
where $[\boldsymbol{\xi}^*\boldsymbol{\Sigma}^{*}(\hat{\boldsymbol{\xi}}^{*})^{\t}]_{r} $ denotes the $r$-th row of 
the matrix $\boldsymbol{\xi}^*\boldsymbol{\Sigma}^{*}(\hat{\boldsymbol{\xi}}^{*})^{\t}$, 
and $\lambda_r^{*\nu} = [\boldsymbol{{\xi}}^*\boldsymbol{{\Sigma}^{*}}(\boldsymbol{\hat{\xi}^{*}})^{\t}]_{r} \bx^{\nu}/{\sqrt{d}}$ represents the $r$-th element of the teacher local field vector
$\boldsymbol{\lambda}^{* \nu} \in \mathbb{R}^{k}$.
The teacher network is quenched as
$[\boldsymbol{{\xi}}^*\boldsymbol{{\Sigma}^{*}}(\boldsymbol{\hat{\xi}^{*}})^{\t}] _{ij}\sim \mathcal{O}(1)$.
Here, we focus on the non-linear transfer function $\sigma(x)=\operatorname{erf}(x / \sqrt{2})$. In addition,
we train the other shallow network called the student network, by minimizing the loss function
$\mathcal{L}(y, \hat{f}(\bx, \Theta))$ over the training data (labels are given by the teacher network),
where $\Theta$ denotes the trainable parameters. The student's prediction for a fresh sample $\bx$ is given by
\begin{equation}
	\hat{f}(\bx, \boldsymbol{\hat{\xi}},\boldsymbol{{\Sigma}},\boldsymbol{{\xi}})=\frac{1}{m} \sum_{r=1}^{m} \sigma\left(\frac{[\boldsymbol{{\xi}}\boldsymbol{{\Sigma}}(\boldsymbol{\hat{\xi}})^{\t}]_{r} \bx}{\sqrt{d}}\right)=\frac{1}{m} \sum_{r=1}^{m} \sigma\left(\lambda_{r}\right),
\end{equation}
where $\lambda_{r}$ denotes the $r$-th component of the student local field $\boldsymbol{\lambda} =\boldsymbol{{\xi}}\boldsymbol{{\Sigma}}(\boldsymbol{\hat{\xi}})^\top\bx$,
and the student has $m$ hidden neurons. The student is supplied with data samples in sequence (one sample each time step). We next use $\nu$ to indicate the time step as well.

The mean-squared-error can be evaluated as
\begin{equation}
	\ell_{\mathrm{MSE}}(\boldsymbol{\Omega})=\frac{1}{2}\mathbb{E}_{\boldsymbol{\lambda},\boldsymbol{\lambda}^{*} \sim \mathcal{N}\left(\boldsymbol{\lambda}, \boldsymbol{\lambda}^{*} \mid 0, \boldsymbol{\Omega}\right)}\left[\left(\hat{f}(\boldsymbol{\lambda})-f\left(\boldsymbol{\lambda}^{*}\right)\right)^{2}\right],	
\end{equation}
where $f(\cdot)$ indicates the teacher's output, and we have replaced the expectation $\mathbb{E}_{\bx, y \sim \mathbb{P}(\bx, y)}[\cdot]$ 
by $\mathbb{E}_{\boldsymbol{\lambda}, \boldsymbol{\lambda}^{*} \sim \mathcal{N}\left(\boldsymbol{\lambda}, \boldsymbol{\lambda}^{*} \mid 0, \boldsymbol{\Omega}\right)}[\cdot]$,
because of the central-limit theorem and the i.i.d. setting we consider~\cite{Biehl-1995,Saad-1995,Goldt-2019}. 
The covariance of the local field $\boldsymbol{\Omega}^{\nu} \in \mathbb{R}^{(k+m) \times(k+m)}$ can be specified as follows, 
\begin{equation}
	\boldsymbol{\Omega}^{\nu} \equiv\left[\begin{array}{cc}
		\mathbf{Q}^{\nu} & \mathbf{M}^{\nu} \\
		(\mathbf{M}^{\nu})^{\t} & \mathbf{P}
	\end{array}\right],
\end{equation}
where $\mathbf{Q}^{\nu} \equiv \mathbb{E}_{\bx, y \sim \mathbb{P}(\bx, y)}\left[\boldsymbol{\lambda}^{\nu} (\boldsymbol{\lambda}^{\nu})^{ \t}\right]$, 
$\mathbf{M}^{\nu} \equiv \mathbb{E}_{\bx, y \sim \mathbb{P}(\bx, y)}\left[\boldsymbol{\lambda}^{\nu} (\boldsymbol{\lambda}^{* \nu})^{ \t}\right]$,
and $\mathbf{P}^{\nu} \equiv \mathbb{E}_{\bx, y \sim \mathbb{P}(\bx, y)}\left[\boldsymbol{\lambda}^{* \nu} (\boldsymbol{\lambda}^{* \nu})^{ \t}\right]$.
By definition, $\mathbf{P}$ is fixed, while $\mathbf{Q}^{\nu}$ and $\mathbf{M}^{\nu}$ evolve according to the gradient updates, 
following a set of deterministic ordinary differential equations (ODEs) as the input dimension $d\to \infty$~\cite{SM}. These matrices are exactly the order parameters in physics.
For simplicity, we consider $\boldsymbol{\xi} = \boldsymbol{\xi}^*$ and $\boldsymbol{\Sigma} = \boldsymbol{\Sigma}^*$, i.e., only the upstream patterns are learned.
\begin{figure}
	\centering
	\includegraphics[bb=1 1 503 553,width=0.5\textwidth]{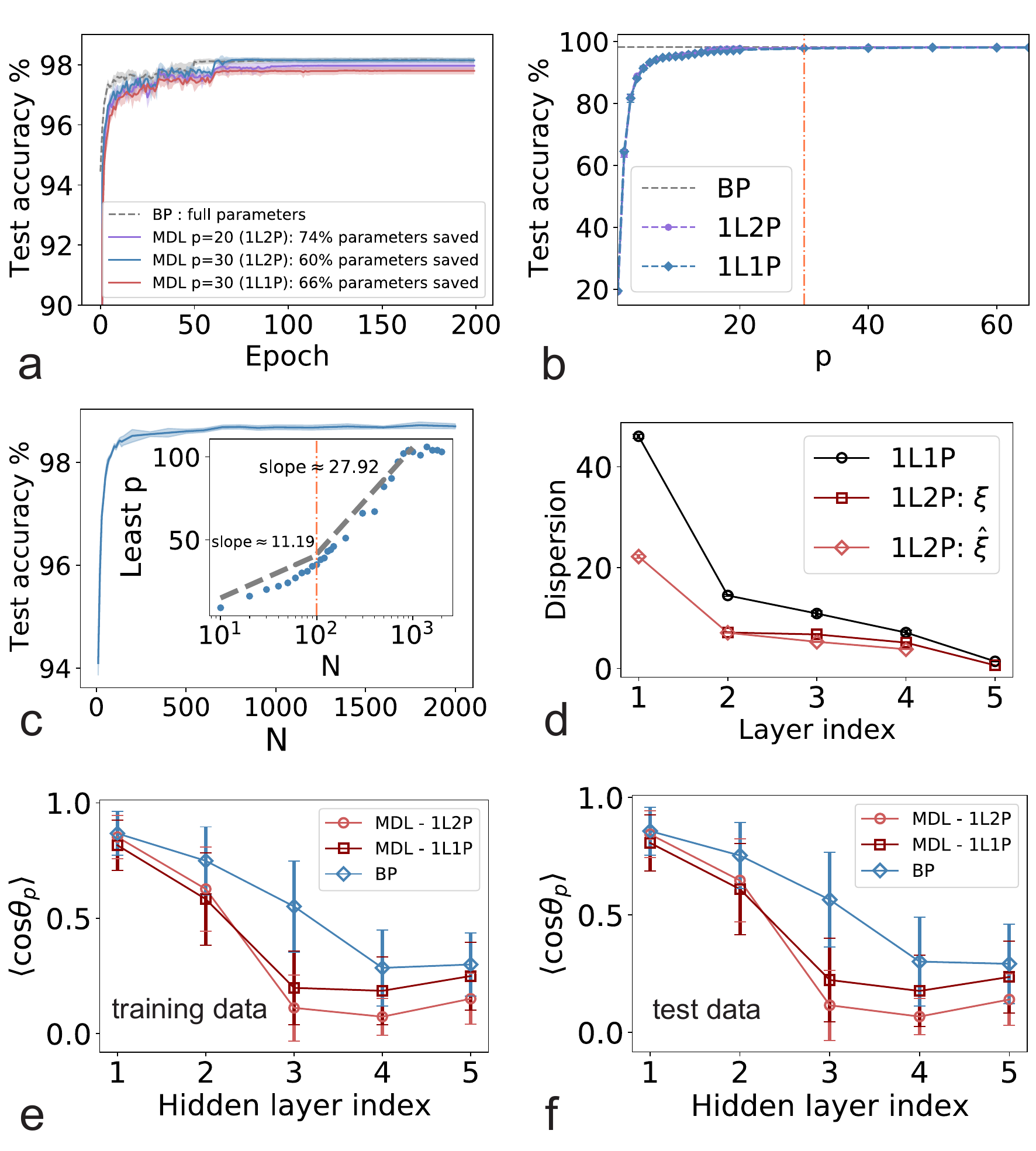}
	\caption{Test performance and mode hierarchy of MDL in deep neural networks. 
	(a) Training trajectories of a four-layer network, indicated by $784$-$100$-$100$-$10$, where each number indicates the 
	corresponding layer width. The number of modes $p^l = p$ for layer $l$, where $l = 1,\ldots,L$. 
	$p = 20$, or $p=30$.
	Networks are trained on the full MNIST dataset ($6\times 10^4$ images) and tested on an unseen dataset containing $10^4$ images. 
	The fluctuation is computed over five independent runs. 
	(b) Testing accuracy versus $p$ (the number of modes is the same for all layers).
	The same architecture as (a) is used. The error bar characterizes the fluctuation across
	five independently trained networks, and each marker denotes the average result. 
	The least number of modes is indicated by the dash-dot line.
	 (c) The performance changes with the network width. The inset shows the least number of modes versus the layer width $N$ (in the logarithmic scale).
	 The network architecture is given by $784$-$N$-$N$-$10$. 
	 The dash-dot line in the inset separates the piecewise logarithmic increase ($\propto\ln N$) regions.
	The result is
	obtained from five independent runs.
	(d) The averaged Euclidean distance (dispersion) from the pattern-cloud center 
	$(\frac{1}{p}\sum_{\alpha}\boldsymbol{\xi}^l_{\alpha})$ as a function of layer index. 
	 The network architecture is specified by $784$-$100$-$100$-$100$-$10$ ($p=30$).
	 (e-f) Subspace overlap (principal angle) versus layer. The overlap is averaged with five independent runs, and seven-layer networks with hidden-layer width $100$ are trained ($p=30$).
	}\label{fig2}
\end{figure}

\textit{Results.}---
MDL can reach a similar test accuracy with that of BP performed
in the weight space, when $p$ is sufficiently large [Fig.~\ref{fig2} (a)].
The computational cost of the BP scales with $N_l^2$. In contrast, MDL works in the mode space, requiring a training cost of only the order of
$pN_l$. Note that $p$ is much smaller than $N_l$ (or $\lim_{N_l\to\infty}p^l/N_l=0$), and our MDL does not need any additional training constraints (compared to other matrix factorization algorithms~\cite{SM}).
Remarkably, when $p=30$, the performance of MDL already matches that of BP [Fig.~\ref{fig2} (b)], but only utilizes $40\%$ of the full sets of parameters that are consumed by the BP.
In fact, each hidden layer can have two different types of latent pattern (1L2P) due to the mode decomposition. But if we assume that
$\boldsymbol{{\xi}}^{l} = \boldsymbol{\hat{\xi}}^{l}$, i.e., each layer share a single type of pattern (1L1P), we can further reduce the computational cost by an amount of $\sum_lp^lN_l$, without
sacrificing the test accuracy [Fig.~\ref{fig2} (b)]. Varying the network width, we reveal a {\it logarithmic} increase of the least number of modes [Fig.~\ref{fig2} (c)],
which is a novel property of deep learning in the mode space, in stark contrast to a linear number of memory patterns in previous studies~\cite{Jiang-2021a}. When the network width further grows,
the least number can even become a constant. We argue that this manifests three separated phases of poor-good-saturated performance with increasing layer width (see Fig. S9 in~\cite{SM}).

\begin{figure}
	\centering
	\includegraphics[bb=2 2 817 272,width=0.5\textwidth]{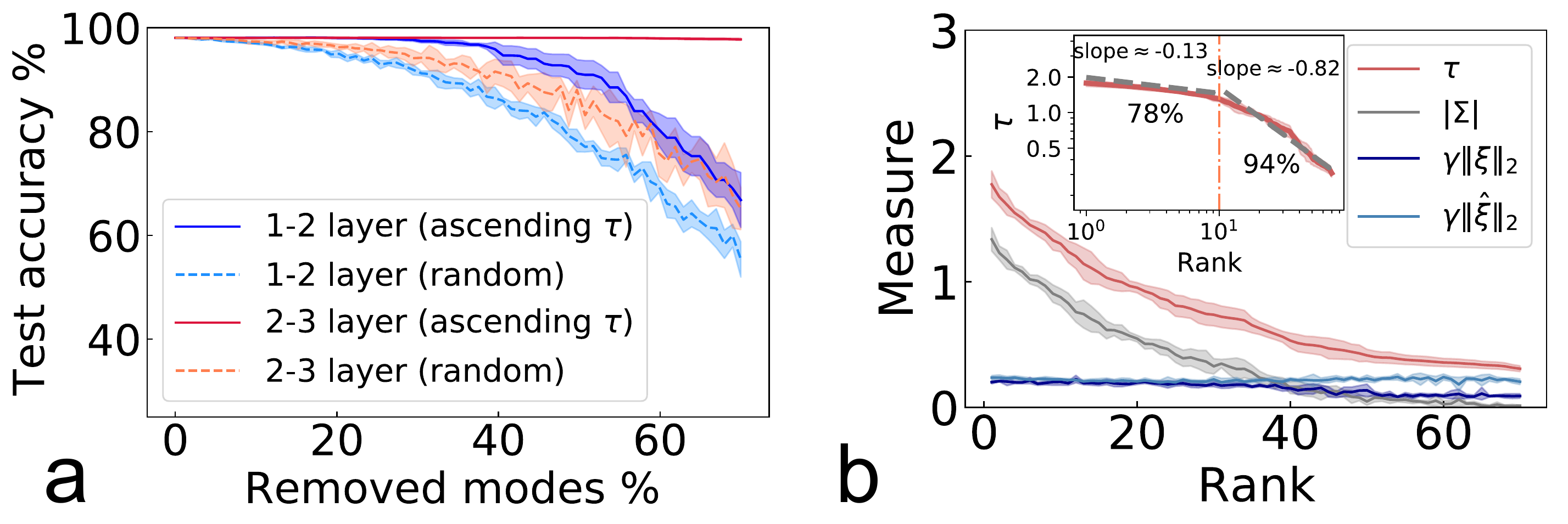}
	\caption{The robustness properties of well-trained four-layer
	MDL models with the architecture $784$-$100$-$100$-$10$. The case of 1L1P is considered with $p=70$ in the hidden layers. 
	(a) Effects of removing modes through two protocols: removing modes with weak measure $\tau$
	first (solid line) and removing modes randomly (dashed line).  
	The fluctuation is computed over ten independent runs. 
	(b) The rescaled $\ell_2$ norms $\gamma\Vert \xi\Vert_2$, $\gamma\Vert \hat{\xi}\Vert_2$ and 
	the absolute values of $\Sigma$ versus their rank (in descending order) in 
	the hidden layers, where
	$\gamma = \sum_{\alpha}|{\Sigma_\alpha}|/\sum_{\alpha}(\Vert {\xi}_\alpha\Vert_2+\Vert {\hat{\xi}}_\alpha\Vert_2)$. 
	The inset shows a log-log plot of the $\tau$ measure, displaying a piecewise power-law behavior.
	The error bar is computed over five independent runs. The marked percentage indicates
	the generalization accuracy after removing the corresponding side of modes. 
	}\label{fig3}
\end{figure}

To see how the latent patterns are transformed in geometry along the network hierarchy, we first calculate the center of the pattern space.
Then the Euclidean distance from this center to each pattern is analyzed.
We find that the pattern space becomes progressively compact when going to deep layers [Fig.~\ref{fig2} (d)].
To further characterize the geometric details, we define the subspace spanned by the principal eigenvectors of the layer neural responses to one type of inputs.
Then the subspace overlap is calculated as the cosine of the principal angle between two subspaces corresponding to two types of inputs~\cite{Bjo-1973,SM}.
We find that
the hidden-layer representation becomes more disentangled with layer in comparison with BP [Fig.~\ref{fig2} (e,f)].
MDL shows great computational benefits of representation disentanglement, thereby facilitating discrimination. A slight increase of the overlap is observed for deeper layers,
which is caused by the saturation of the test performance (see more analyses in~\cite{SM}).

Compared to other matrix factorization methods, MDL has no additional constraints for the modes and importance scores, therefore being flexible for feature extraction. 
We find that the interlayer patterns are less orthogonal than the intralayer ones. The geometric transformation carried out by 
these latent pattern matrices is not strictly a rotation for which the $\ell_2$ norm is preserved. This flexibility may be the key to make our method better than other matrix factorization
methods in both training cost and learning performance (see details in~\cite{SM}).

\begin{figure}
	\centering
	\includegraphics[bb=3 2 843 281,width=0.5\textwidth]{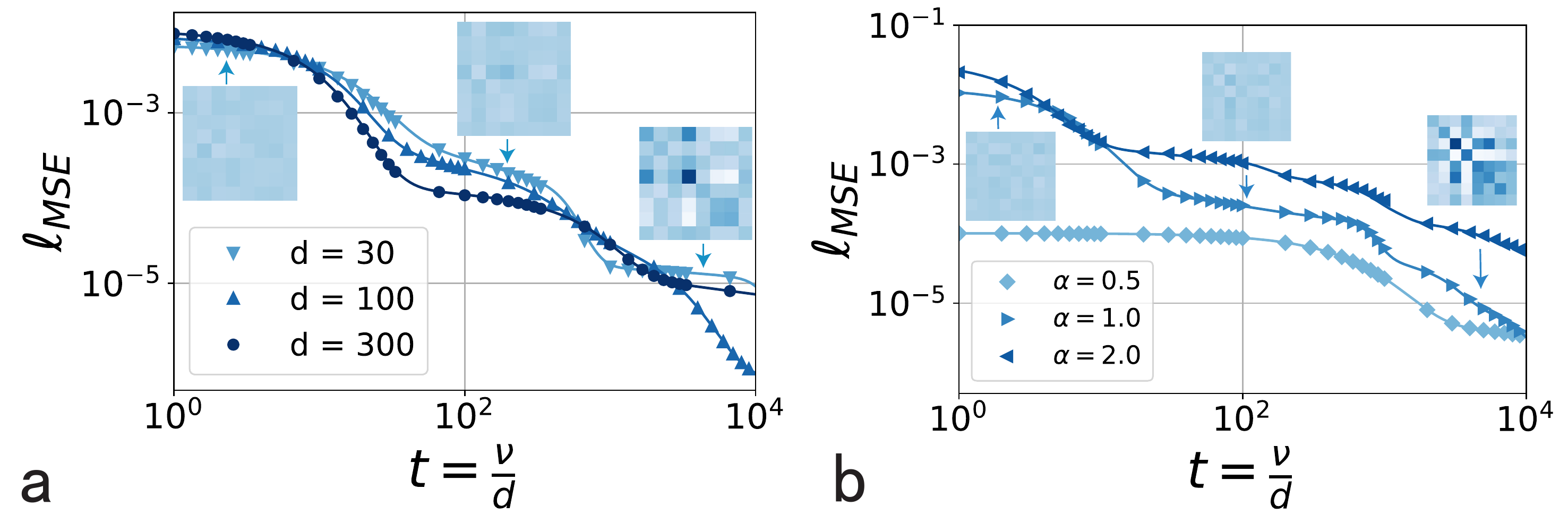}
	\caption{Mean-squared error dynamics in terms of $t=\frac{\nu}{d}$, 
	where $\nu$ denotes the on-line sample index, and $d$ is the input dimension.
	The teacher and student networks share the same number of hidden neurons ($m=k=8$).
	Markers represent results of the simulation, while the solid lines denote the
	theoretical predictions from solving the mean-field ODEs. The number of modes $p^*=p=\alpha\ln d$ ($\alpha$ 
	denotes the mode load here).
	(a) Fixed $\alpha=1$. 
	 (b) Fixed $d=100$. The color deepens as $\alpha$ or $d$ increases. The insets display the evolving $\mathbf{M}$ matrix for $d=30$ and $\alpha=1.0$, respectively.
	}\label{fig4}
\end{figure}

We next ask whether some modes are more important than the others. 
Therefore, we rank the modes according to the
measure $\tau_\alpha = \gamma\Vert \xi_\alpha\Vert_2+\gamma\Vert \hat{\xi}_\alpha\Vert_2+|\Sigma_\alpha|$, 
where $\gamma = \sum_{\alpha}|{\Sigma_\alpha}|/\sum_{\alpha}(\Vert {\xi}_\alpha\Vert_2+\Vert {\hat{\xi}}_\alpha\Vert_2)$ 
to make comparable the magnitudes
of the pattern and importance ($\boldsymbol\Sigma$) score. 
Removing modes with weak values of $\tau$ first yields much higher accuracy than the random removal protocol [Fig.~\ref{fig3} (a)], 
suggesting the existence of leading modes. Moreover, deeper layers are more robust. Figure~\ref{fig3} (b) shows the measure as a function of rank in descending order, which can be
approximately captured by piecewise power-law behavior (a transition point at the rank $10$). Ranking with only the importance scores yields similar behavior~\cite{SM}.
A small exponent is observed for the leading measures, while the remaining measures bear a large exponent, thereby 
revealing the coding hierarchy of latent modes in the deep networks. This intriguing behavior does not change with the regularization strength or the hidden-layer width~\cite{SM}.

Finally, the on-line mean-squared error dynamics of our model can be predicted perfectly in a teacher-student setting.
The number of modes strongly affects the shape of the learning dynamics, and a large mode load can make the plateaus disappear (Fig.~\ref{fig4}).
Moreover, during learning, the alignment between receptive fields of the student's hidden nodes and the teacher's ones continuously emerge, which is called the specialization transition~\cite{Sch-1993,Goldt-2019}. 

\textit{Conclusion.}---
In this Letter, we propose a mode decomposition learning that works in the mode space rather than the conventional weight space.
This learning scheme has three-fold technical and conceptual advances. First, the learning can achieve the comparable performance with standard methods, with a significant reduction of training costs.
We also find that the least number of modes grows only logarithmically with the network width and becomes even independent of larger width, which is in stark contrast to a linear number of patterns in recurrent memory networks.
Second, the learning leads to progressively compact pattern spaces, which promotes highly disentangled hierarchical representations.
The upstream pattern maps the activity into a low-dimensional space, and then the resulting embedding is 
further expanded or contracted. After that, the modified embedding is re-mapped into the high-dimensional activity space. This sequence of geometric transformation
can be understood as a linear-nonlinear hidden structure. Third, all modes are not equally important to the generalization ability of the network, showing an intriguing
piecewise power-law behavior. Finally, the mode learning dynamics can be predicted by the mean-field ODEs, revealing the mode specialization transition.
Therefore, the MDL inspires a rethinking of conventional deep learning, offering a faster, more interpretable training framework. Future works along this direction will be inspired.
For example, the impact of other structured dataset, mode dynamics in over-parameterized or recurrent networks, and the origin of adversarial vulnerability of deep networks in terms of geometry of the mode space.

\section{Acknowledgments}
This research was supported by the National Natural Science Foundation of China for
Grant number 12122515, and Guangdong Provincial Key Laboratory of Magnetoelectric Physics and Devices (No. 2022B1212010008), and Guangdong Basic and Applied Basic Research Foundation (Grant No. 2023B1515040023).

\setcounter{figure}{0}    
\renewcommand{\thefigure}{S\arabic{figure}}
\renewcommand\theequation{S\arabic{equation}}
\setcounter{equation}{0}  

\newpage
\onecolumngrid
\appendix
\section*{Supplemental Material}
\label{SM}
\section{Derivation of learning equations}
In this section, we show how to derive the updating equations for 
the mode parameters $\boldsymbol{\theta}^l = (\boldsymbol{\hat{\xi}}^l, \boldsymbol{\Sigma}^l, \boldsymbol{{\xi}}^{l+1})$ where
the superscript $l$ indicates the layer index in the range from $1$ to $L$. The loss function 
is the cross entropy $\mathcal{C}=-\sum_{i} \hat{h}_{i} \ln h_{i}$ averaged over all training examples (divided into mini-batches in stochastic gradient descent),
where $\hat{h}_i$ is defined as the target label (one-hot representation as common in machine learning).
After training the network on the training dataset with size $T$, we evaluate the generalization
performance of the network on the unseen dataset with size $V$.

In our framework of mode decomposition learning, the weight is decomposed into the form as follows,
\begin{equation}
	\mathbf{w}^l = \boldsymbol{\hat{\xi}}^l\boldsymbol{\Sigma}^l (\boldsymbol{{\xi}}^{l+1})^{\t}.
\end{equation} 
The mode parameters are updated according to gradient descent of the loss function,
\begin{equation}
	\Delta \boldsymbol{\theta}_{ij}^{l}=-\eta \mathcal{K}_{j}^{l+1} \frac{\partial z_{j}^{l+1}}{\partial \boldsymbol{\theta}_{ij}^{l}},
\end{equation}
where $\eta$ denotes the learning rate, and the error propagation term
$\mathcal{K}_{j}^{l+1} \equiv\partial \mathcal{C} / \partial z_{j}^{l+1}$. On the top layer, $\mathcal{K}_{j}^{l+1}$ can be computed with
the result $\mathcal{K}_j^L = -\hat{h}_{j}^{L}\left(1-h_{j}^{L}\right)$. For lower layers, the term $\mathcal{K}_{i}^{l}$ can be iteratively computed
using the chain rule. More precisely,
\begin{equation}
	\begin{aligned}
	\mathcal{K}_i^l &= \partial \mathcal{C} / \partial z_{i}^{l} = \sum_{j}\frac{\partial \mathcal{C} }{\partial z_{j}^{l+1}}\frac{\partial z_{j}^{l+1}}{\partial z_{i}^{l}}\\
	& =\sum_j \mathcal{K}_j^{l+1}\sum_\a\xi^{l+1}_{i\a}\Sigma^{l}_\a\hat{\xi}_{j\a}^{l}f^{\prime}(z_i^l).
	\end{aligned}
\end{equation}
The explicit expressions of gradient steps for the three sets of mode parameters are given as follows,
\begin{equation}
	\begin{aligned}
		\Delta \xi^{l+1}_{j \a}& = -\eta\frac{\partial \mathcal{C}}{\partial \xi^{l+1}_{j \a}} = -\eta \mathcal{K}_j^{l+1}\sum_{i}\Sigma_{\a}^l\hat{\xi}_{i\a}^{l}h_i^{l},\\
		\Delta \Sigma_{\a}^l& = -\eta\frac{\partial \mathcal{C}}{\partial \Sigma^l_{\a}} = -\eta\sum_j \mathcal{K}_j^{l+1}\sum_i \xi_{j\a}^{l+1} \hat{\xi}_{i \a }^l h_{i}^l,\\
		\Delta \hat{\xi}^l_{i \a } & = -\eta \frac{\partial \mathcal{C}}{\partial\hat{\xi}_{i \a}^{l}} = -\eta \sum_j \mathcal{K}_j^{l+1} \xi_{j \a}^{l+1}\Sigma_{\a}^lh_i^l.\\
	\end{aligned}
\label{eq1}
\end{equation}
The above learning equations apply to the case of 1L2P case.

Next, we consider the 1L1P case ($\boldsymbol{{\xi}}^{l} = \boldsymbol{\hat{\xi}}^{l}$).  
Apart from the single input pattern for the first layer $\boldsymbol{\hat{\xi}}^{1}$ and the single output pattern for 
the last layer $\boldsymbol{\xi}^L$, two types of pattern in each hidden layer $[\boldsymbol{{\xi}}^{l}, \boldsymbol{\hat{\xi}}^{l}]$ take
the same form, and we denote $\boldsymbol{{\xi}}^{l} = \boldsymbol{\hat{\xi}}^{l} = \boldsymbol{\Xi}^l$.
The expression of $\mathcal{K}_i^l$ remains unchanged, and we can then update $[\boldsymbol{\hat{\xi}}^1, \boldsymbol{\Sigma}^l, \boldsymbol{{\xi}}^{L}]$ 
according to Eq.~\eqref{eq1}. Next, we give the gradient descent equation for $\boldsymbol{\Xi}^l$ where $l = 2,..., L-1$ as follows
\begin{equation}
	\begin{aligned}
		\Delta \Xi^l_{j \a}& = -\eta\frac{\partial \mathcal{C}}{\partial \Xi^l_{j \a}} = -\eta \mathcal{K}_j^{l}\sum_{i}\Sigma_{\a}^{l-1}\hat{\xi}_{i\a}^{l-1}h_i^{l-1}-\eta\sum_i\mathcal{K}_i^{l+1}\xi_{i\a}^{l+1}\Sigma_{\a}^{l}h_j^{l},
	\end{aligned}
	\label{eq2}
\end{equation}
where two terms contribute to the gradient---the first one comes from the contribution of $\boldsymbol{\xi}^l$,
while the second one originates from the fact that the same pattern can act as $\boldsymbol{\hat{\xi}}^{l}$.

To ensure the weighted sum in the pre-activation is independent
of the upstream layer width and the number of modes $p^l$, we choose the initialization scheme such that
$[\boldsymbol{{\xi}}^{l+1}\boldsymbol{{\Sigma}}^{l}(\boldsymbol{\hat{\xi}}^{l})^{\t}]_{ij}\sim \mathcal{O}(\frac{1}{\sqrt{N_l}})$. This scaling is inspired by
studies of Hopfield models~\cite{Jiang-2021a}.
In practice, we independently and identically sample the initial elements $\xi^{l+1}_{i\a},\Sigma^l_{\a},\hat{\xi}^l_{j\a}$ 
from the standard Gaussian distribution, and then the weight values are multiplied by a factor of $1/\sqrt{N_l\ln N_l}$.
Note that the number of modes are assumed to be proportional to $\ln N_l$. But if the number is a constant denoted by $P^l$, then the factor could
be $1/\sqrt{P^lN_l}$.
\begin{figure}[h]
\centering
\includegraphics[bb=4 6 402 274,scale = 0.6]{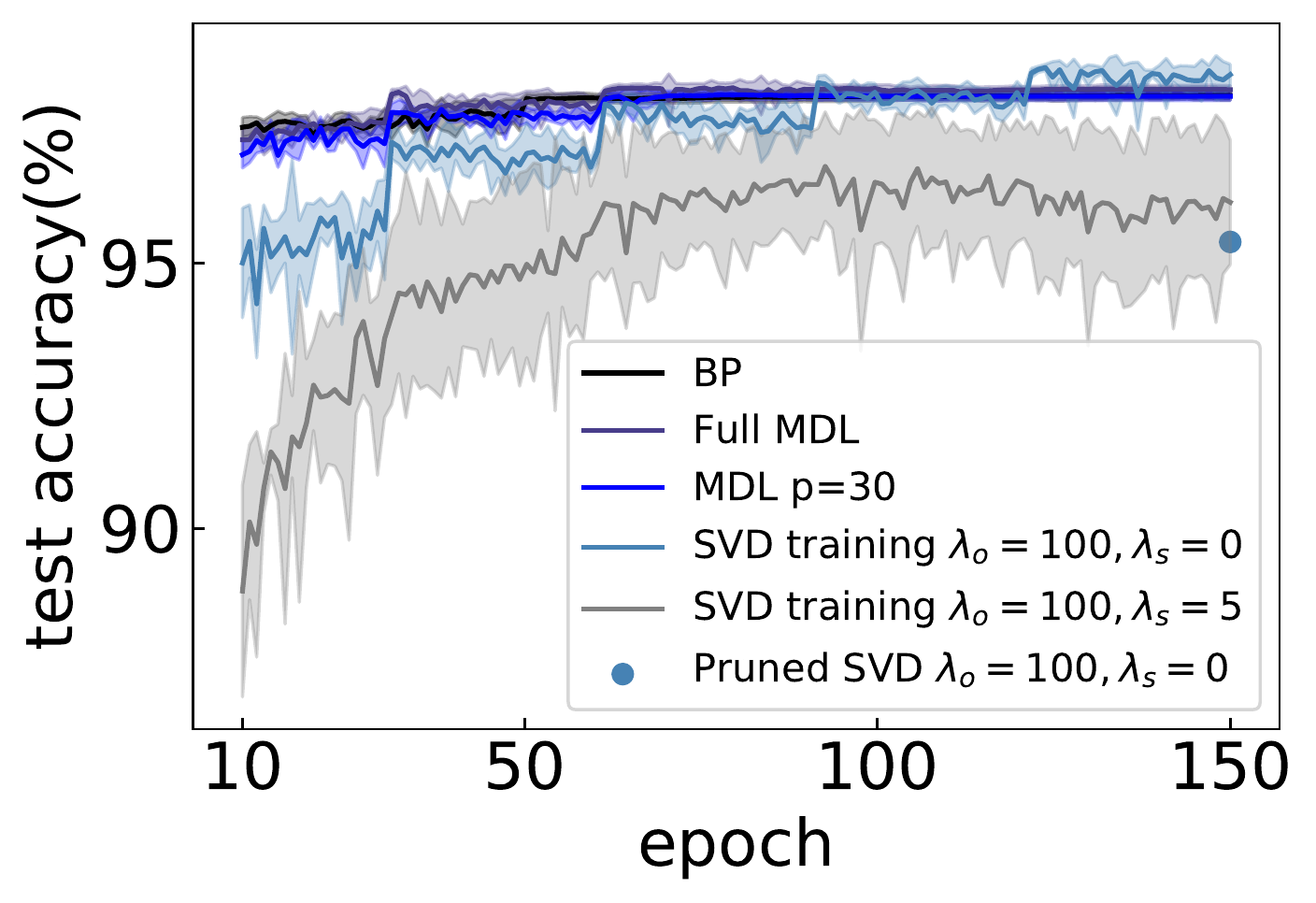}
\caption{The comparison among the SVD training, MDL (1L2P) and traditional BP in learning performance.
The network structure is specified by $[784,100,100,10]$ in all cases. The full MDL indicates the MDL with the same number of parameters as that of the SVD training,
while the blue dot (pruned SVD) indicates the pruning of the full SVD model $60\%$ (the modes with small $|s_i|$ ranked in descending order) modes off each layer (except the output layer) to make the consuming parameter amount
comparable with that of MDL with $p=30$.}
\label{figS2}
\end{figure} 

\begin{figure}[h]
\centering
\includegraphics[bb=2 5 415 275,scale = 0.6]{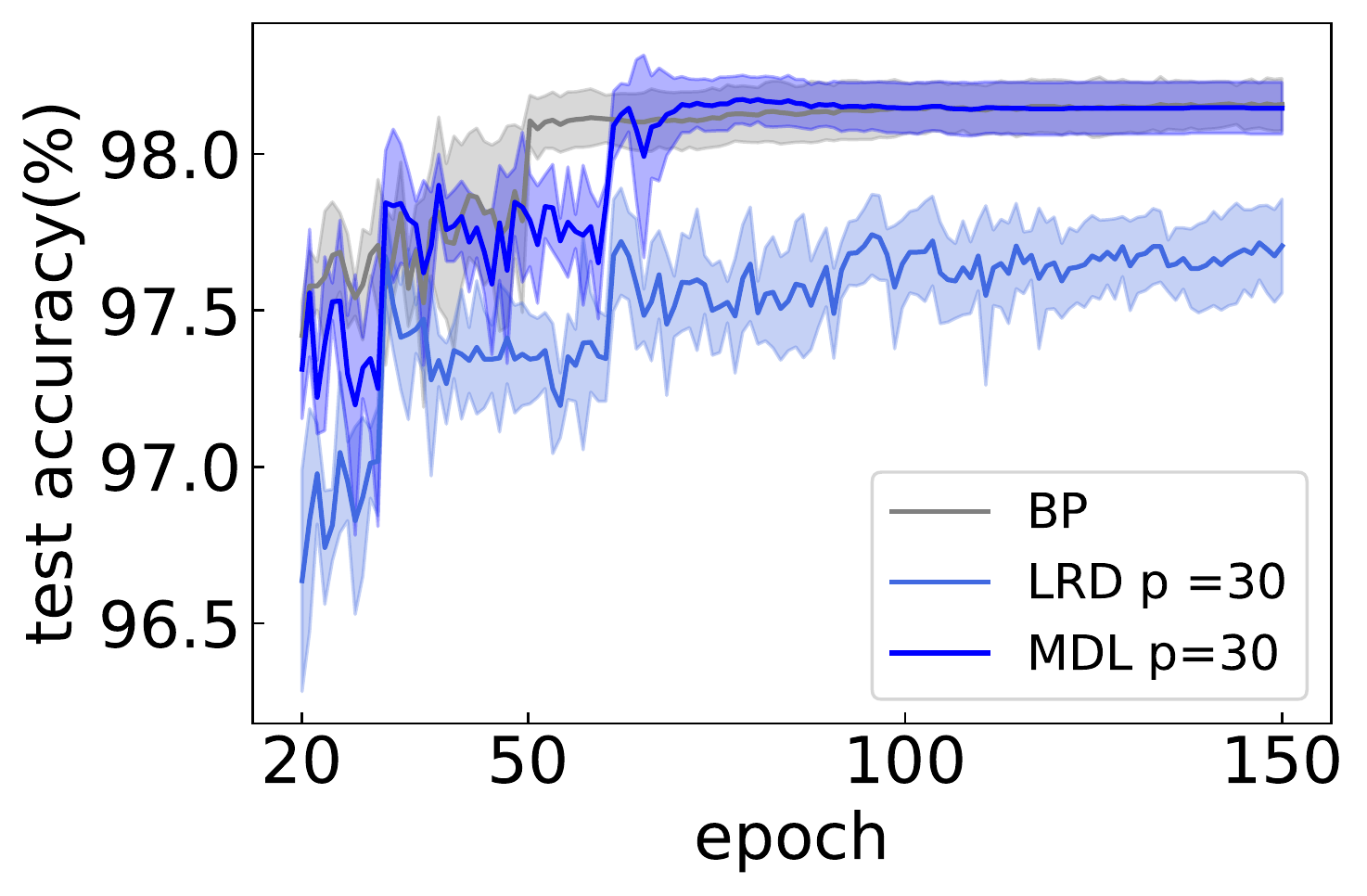}
\caption{The comparison among low-rank decomposition (LRD), MDL (1L2P) and traditional BP. 
Both decomposition methods use $p=30$. The network structure is $[784,100,100,10]$ in all cases.}
\label{figS3}
\end{figure} 

\section{Comparison to other matrix factorization methods}
Here, we compared our MDL method to other matrix factorization methods in learning performance. These other methods include singular value decomposition (SVD),
low rank decomposition (LRD) and spectral training~\cite{SVD-2020,PRE-2021}.

First, the SVD learning scheme is implemented by decomposing the weight of each layer as 
\begin{equation}
\boldsymbol{W}^l=\boldsymbol{U}^l \operatorname{diag}(\boldsymbol{s}^l) (\boldsymbol{V}^l)^\top,
\end{equation}
where the diagonal matrix contains $\min(N_l,N_{l+1})$ non-zero elements in the diagonal, and the elements of $\mathbf{s}^l$ is constrained to be positive.
The orthogonality is forced by two regularization terms as
\begin{equation}
L(\boldsymbol{U}, \boldsymbol{s}, \boldsymbol{V})=  L_T
+\lambda_o \sum_{l=1}^D L_o\left(\boldsymbol{U}_l, \boldsymbol{V}_l\right)+\lambda_s \sum_{l=1}^D L_s\left(\boldsymbol{s}_l\right),
\end{equation}
where $L_T$ is the original training loss, $L_o(\boldsymbol{U}, \boldsymbol{V})=\frac{1}{r^2}\left(\left\|\boldsymbol{U}^T \boldsymbol{U}-\boldsymbol{I}\right\|_F^2+\left\|\boldsymbol{V}^T \boldsymbol{V}-\boldsymbol{I}\right\|_F^2\right)$,
and $L_s(\boldsymbol{s})=\frac{\|\boldsymbol{s}\|_1}{\|\boldsymbol{s}\|_2}=\frac{\sum_i\left|s_i\right|}{\sqrt{\sum_i s_i^2}}$. 
$r$ is the rank of $\boldsymbol{U}$ and $\boldsymbol{V}$, $\|\bullet\|_F$ denotes the Frobenius norm of a matrix.
The regularization term $L_o$ forces $\boldsymbol{U}$ and $\boldsymbol{V}$ to be orthogonal, while $L_s$ adjusts the sparsity level of $\boldsymbol{s}$.
The gradients for each set of parameters are derived below,
\begin{equation}
\begin{aligned}
\frac{\partial L_o}{\partial \boldsymbol{U}} &= \frac{4}{r^2}\left(\boldsymbol{U}^\top \boldsymbol{U}-\boldsymbol{I}\right)^\top\times \boldsymbol{U}^\top,\\
\frac{\partial L_o}{\partial \boldsymbol{V}} &= \frac{4}{r^2}\left(\boldsymbol{V}^\top \boldsymbol{V}-\boldsymbol{I}\right)^\top\times \boldsymbol{V}^\top,\\
\frac{\partial L_s}{\partial s_i} &=\frac{\operatorname{sign}(s_i)\sqrt{\sum_i s_i^2} - \sum_i |s_i|(\sum_i s_i^2)^{-\frac{1}{2}}s_i}{\sum_i s_i^2}.
\end{aligned}
\end{equation}

For comparison, we carried out the SVD learning, with $L_o = 100$, $L_s = 0.0$, and  $L_o = 100$, $L_s = 5.0$, as shown in Fig.~\ref{figS2}.
We remark that the training cost is larger for SVD models,
which can be calculated as $\sum_l [N_l\times N_{l+1} + \min{(N_l, N_{l+1})}^2+\min{(N_l, N_{l+1})}]$. Taking $[784,100,100,10]$ as
an example, the learning needs $109710$ parameters in total. However, for our MDL with $p=30$ which already reaches the traditional BP performance, 
the learning only needs $35910$ parameters (but traditional BP needs $89400$ parameters).  
In simulations, we prune the full SVD model $60\%$ ((the modes with small $|s_i|$ ranked in descending order)) modes off each layer (except the output layer) to make the number of trainable parameters comparable with that of
MDL with $p=30$. We conclude that the MDL consumes less parameters, yet produces rapid learning with even better performances.

\begin{figure}[h]
\centering
\includegraphics[bb=2 5 415 275,scale = 0.6]{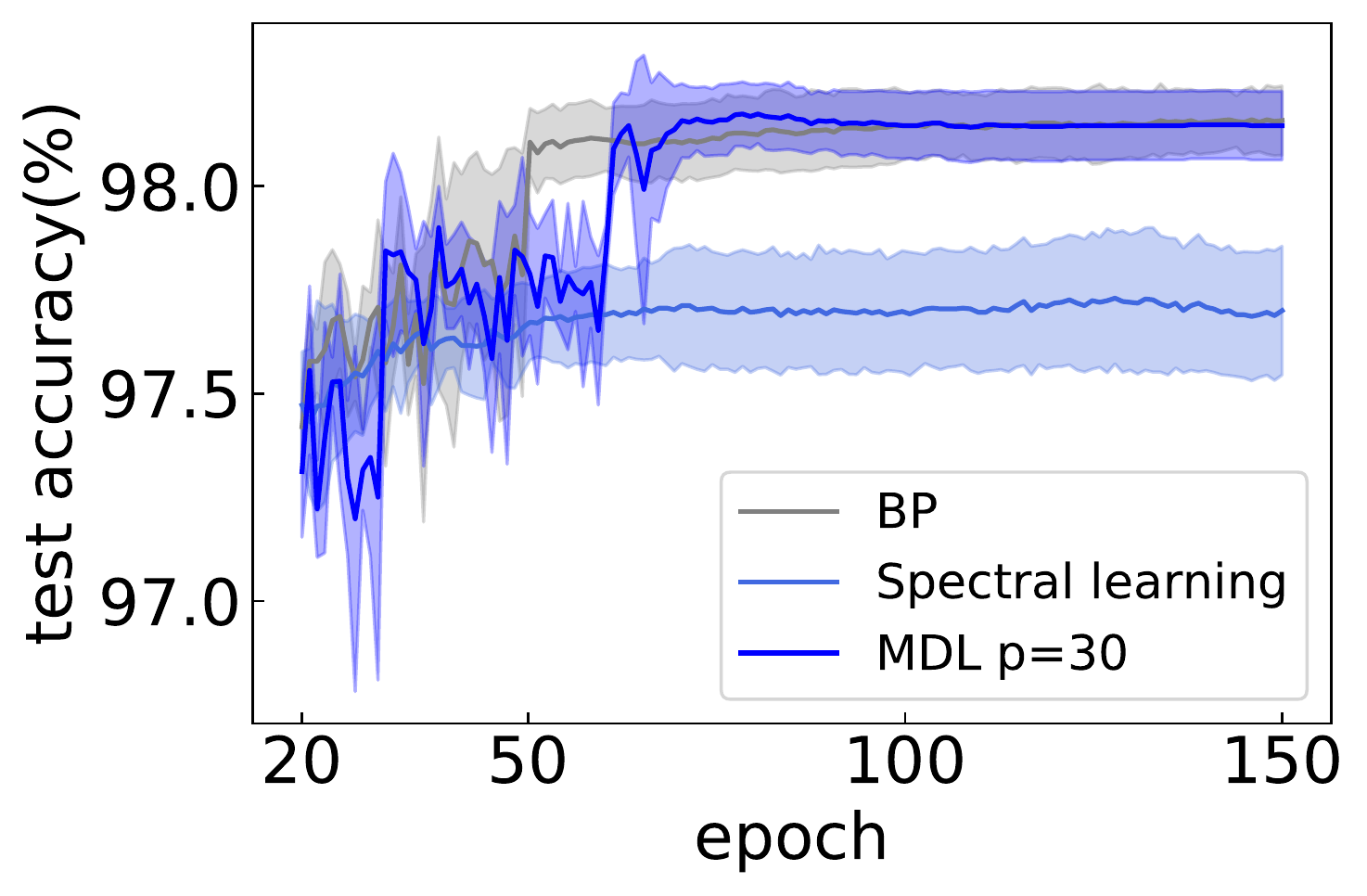}
\caption{The comparison among the spectral learning, MDL (1L2P) and traditional BP. The network structure is $[784,100,100,10]$ in all cases.}
\label{figS4}
\end{figure} 

Next, we fix $\boldsymbol{\Sigma} = \mathbb{I}$ in our MDL, and this reduced form is called the low rank decomposition as follows,
\begin{equation}
\mathbf{W}^{l} = \boldsymbol{\hat{\xi}}^l (\boldsymbol{\xi^{l+1}})^\top.
\end{equation}
In the simulation, we set $p=30$. We can see in Fig.~\ref{figS3} that the performance of the LRD is much worse than that of MDL and traditional BP.

For the recently proposed spectral learning~\cite{PRE-2021}, a carefully-designed transformation matrix $\mathbf{A}^{k}$ (an $N\times N$ matrix, $N$ is the total number of units in the network, and
$k$ is a layer index) is used with a spectral decomposition. The eigenvalues and the associated basis are optimized. However, this training performs worse compared to our MDL in the examples shown in Fig.~\ref{figS4}.

\section{Ranking the modes according to the importance matrix}
\begin{figure}[h]
\centering
\includegraphics[bb=1 2 816 272,scale = 0.6]{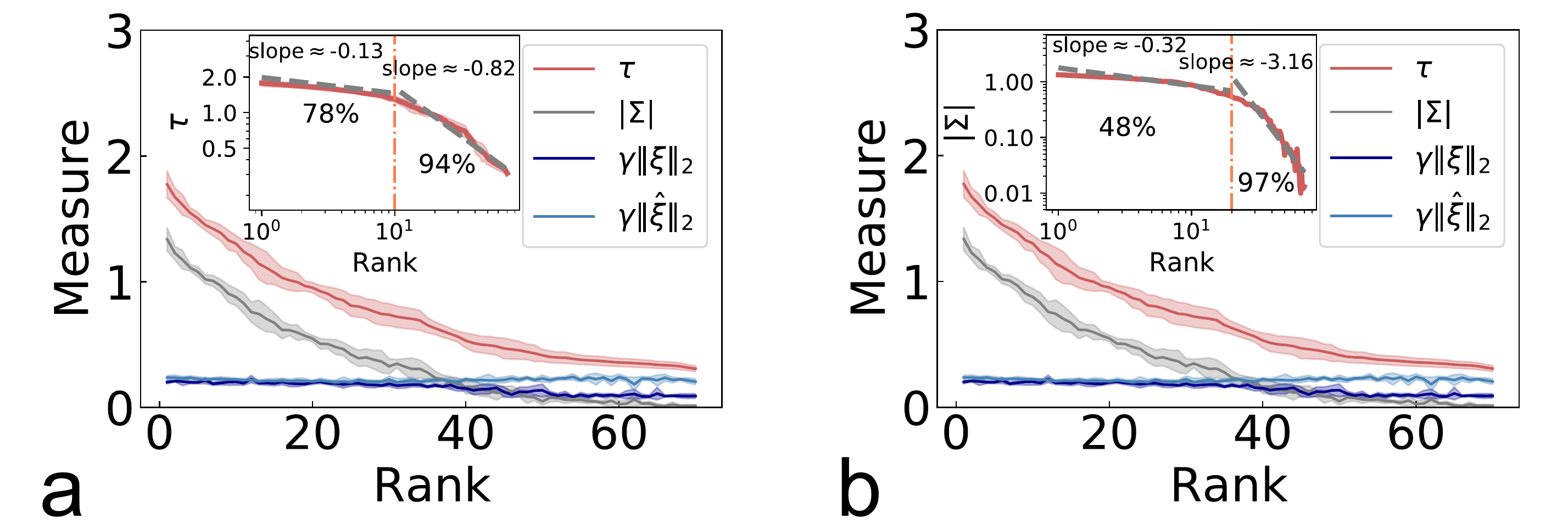}
\caption{Ranking modes. The network structure is $[784, 100, 100, 10]$, and we analyze the 1L1P case here with $p=70$.
The marked percentage in the inset indicates the generalization accuracy after removing the corresponding side of modes in the hidden layer. 
The piecewise power law behavior is retained for both types of ranking.}
\label{figS5}
\end{figure}

Here, we rank the modes according to the diagonal of the importance matrix, rather than the $\tau$ measure. We found that these two ranking schemes lead to qualitatively identical results.
Removing the most important modes (according to either the $\tau$ measure or the importance score) will significantly impair the generalization ability of the network.
Details are illustrated in Fig.~\ref{figS5}. The non-smooth behavior can be attributed to the existence of mode-contribution gap, i.e., the most important modes ($<15\%$ for the $\tau$ measure; $<30\%$ for 
the $\Sigma$ measure) dominate the generalization capability of the network, while other modes capture irrelevant noise in the data.

\section{The qualitative behavior of the MDL does not change with the regularization strength or the hidden-layer width
}
\begin{figure}[h]
\centering
\includegraphics[bb=1 2 816 272,scale = 0.6]{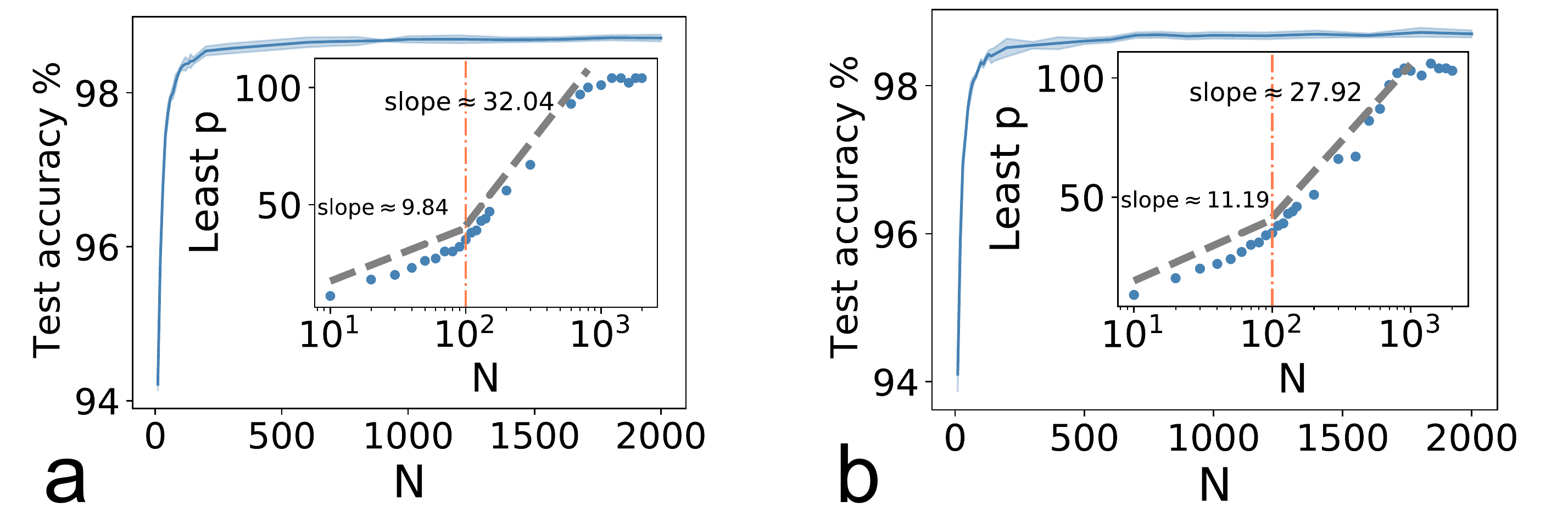}
\caption{The piecewise increasing behavior of the least $p$ with the hidden-layer width.
The network has structure $[784, N, N, 10]$, and we vary $N$ to get the corresponding least $p$, 
which is defined as the least number of modes that MDL needs to reach the performance of the traditional BP. 1L2P case is considered.
(a) and (b) are obtained under different regularization strengths. (a) $\lambda=10^{-3}$.  (b) $\lambda=10^{-4}$. }
\label{figS6}
\end{figure}

\begin{figure}[h]
\centering
\includegraphics[bb=2 3 1249 326,scale = 0.4]{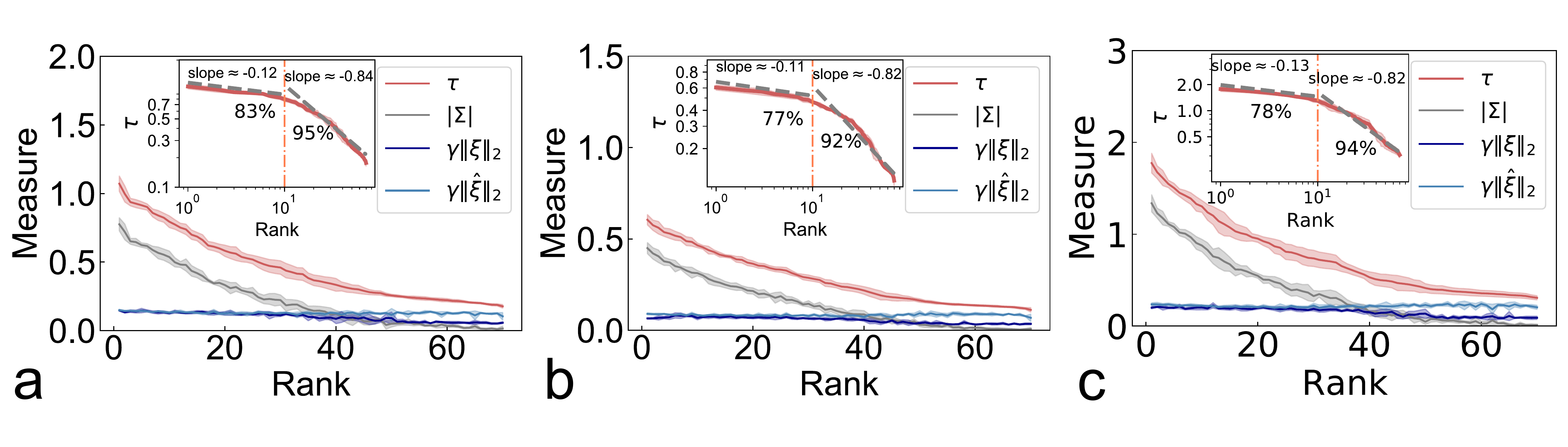}
\caption{The piecewise power law behavior of the $\tau$ measure does not change with the regularization strength $\lambda$ in the 1L1P case.
The network structure is specified by $[784,100,100,10]$. (a) $\lambda=0.01$. (b) $\lambda=0.001$. (c) $\lambda=0.0001$.}
\label{figS7}
\end{figure}

\begin{figure}[h]
\centering
\includegraphics[bb=3 2 566 420,scale = 0.6]{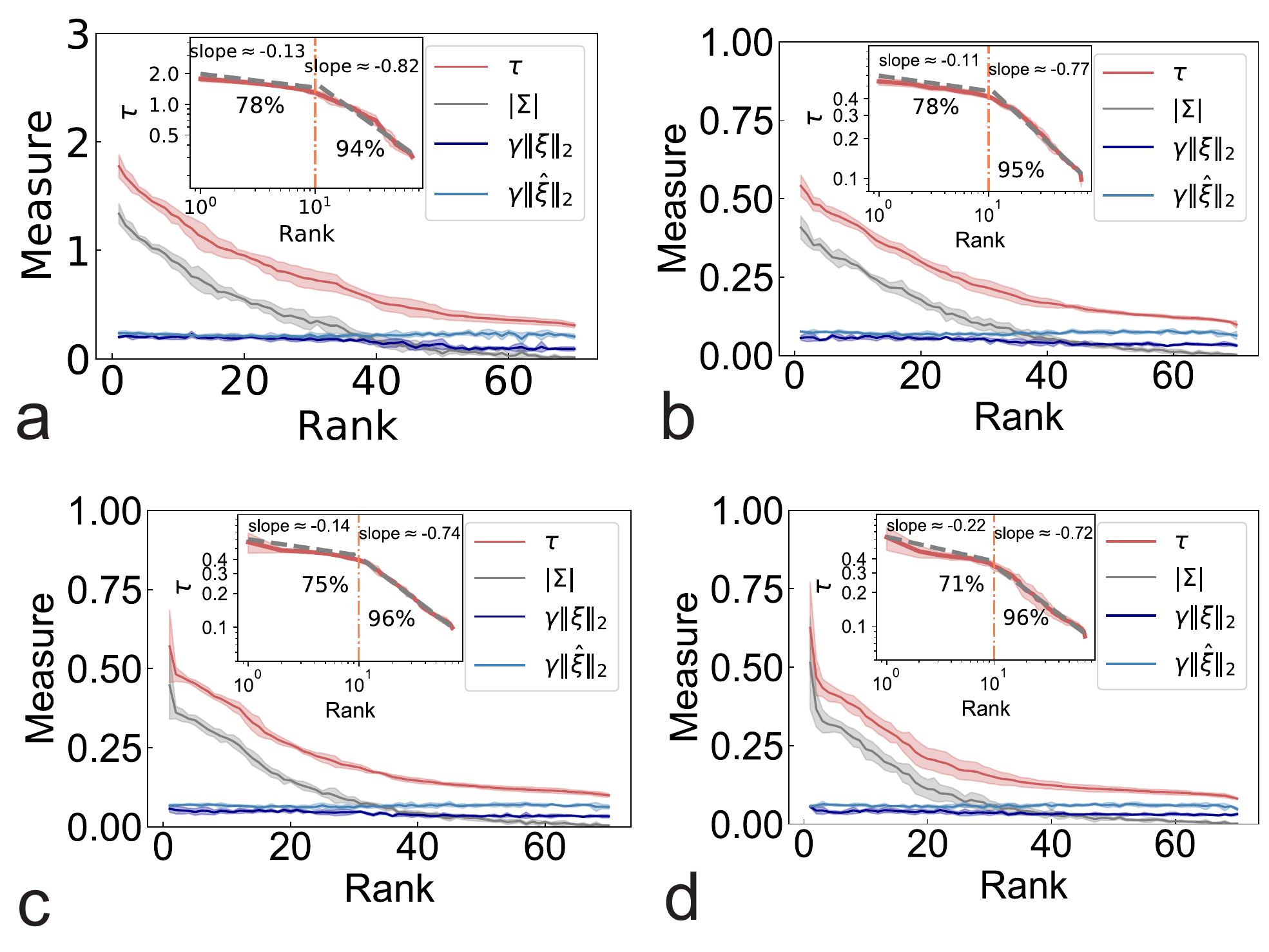}
\caption{The piecewise power law behavior of the $\tau$ measure does not change with the hidden-layer width in the 1L1P case. The network structure is specified by $[784,N,N,10]$.
(a) $N=100$. (b) $N=150$. (c) $N=200$. (d) $N=300$.
} 
\label{figS8}
\end{figure}

\begin{figure}[h]
\centering
\includegraphics[bb=6 37 1262 317,scale = 0.4]{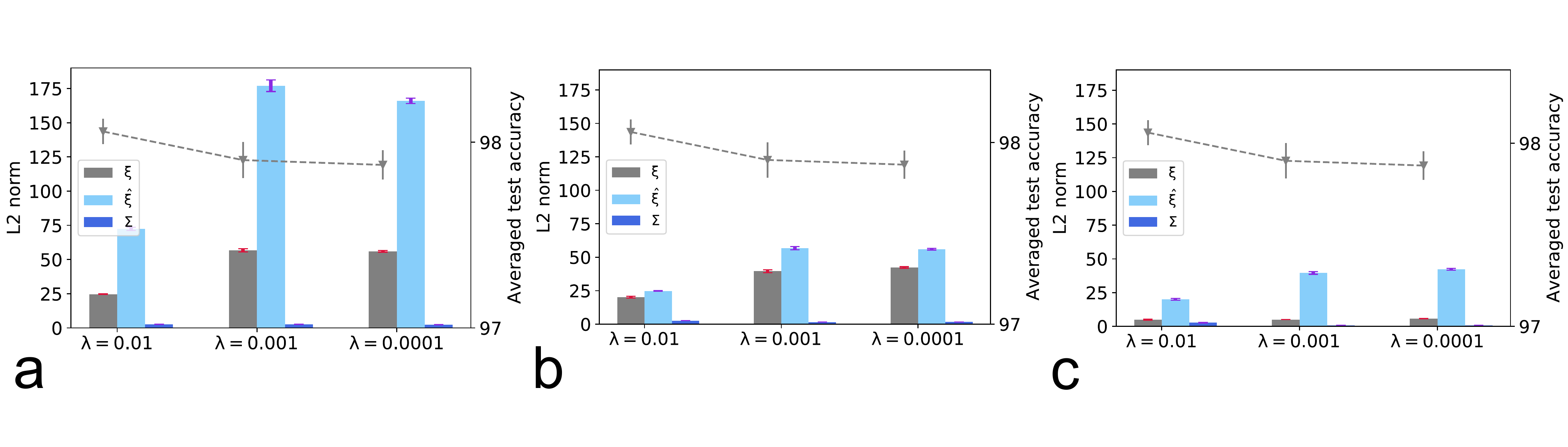}
\caption{The $\ell_2$ norm of parameters $[\boldsymbol{\xi}, \boldsymbol{\hat{\xi}}, \boldsymbol{\Sigma}]$ under three regularization strengths--- $\lambda = 10^{-2},
10^{-3}, 10^{-4}$. The network structure is specified by $[784,100,100,10]$ and $p=70$ for all layers, where the 1L1P case is considered (the 1L2P case yields
qualitatively the same results). (a, b, c) are plotted for the first layer, the second layer, and the output layer, respectively.}
\label{figS9}
\end{figure}

Further, our MDL is in essence a matrix factorization. Therefore, the pattern and importance matrices are not unique. However, in practice, we impose the $\ell_2$ norm level for these patterns and importance scores.
In fact, we find the intriguing properties of the MDL in deep learning do not change with the regularization strength of the $\ell_2$ norm (denoted as $\lambda$, see Fig.~\ref{figS9}). Figure~\ref{figS6} shows an example 
for the behavior of the optimal number of modes versus hidden-layer width, while Fig.~\ref{figS7} shows that the piecewise power law behavior of the $\tau$ measure does not change with the regularization strength.
In addition, the piecewise power law behavior of the $\tau$ measure does not change with the hidden-layer width as well (Fig.~\ref{figS8}).

\section{Subspace overlap of layered response to pairs of stimuli}
In this section, we provide details of estimating the average degree of correlation between neural
responses $\mathbf{h}^l$ to pairs of different input stimuli (e.g., one stimulus contains images of the same class). The covariance of neural response in each layer to the stimulus can be diagonalized to specify a low-dimensional subspace.
The subspace is spanned by the first $K$ principal components. The subspace 
overlap can then be evaluated via the cosine of the principal angle between these two subspaces corresponding to two different stimuli.
In practice, for neural responses in each layer
to the stimulus $s_1$ (e.g., many images of digit $0$), we first identify the first $K$ principal
components of the covariance of $\mathbf{h}_1^\ell$, which explains over $80\%$ of the total variance, and then reorganize the eigenvectors to
an $N_\ell\times K$ matrix, namely $\mathbf{Q}^\ell(s_1)$. We repeat this procedure for another stimulus $s_2$, and get another matrix $\mathbf{Q}^\ell(s_2)$.
Therefore, the columns of $\mathbf{Q}^\ell(s_1)$ and $\mathbf{Q}^\ell(s_2)$ span two subspaces corresponding to the neural 
responses to $s_1$ and $s_2$ respectively. The cosine of the principal angle between these two subspaces is calculated as follows~\cite{Bjo-1973}
\begin{equation}
	\cos \theta_p\left(s_1, s_2\right)=\sigma_{\max }\left({\mathbf{Q}^\ell(s_1)}^{\rm T} \mathbf{Q}^\ell(s_2)\right),
\end{equation}
where $\sigma_{\max}(\mathbf{B})$ denotes the largest singular value of the matrix $\mathbf{B}$.
In simulations, we consider the classification task of the MNIST dataset, where ten classes of digits are fed into
a seven-layer neural network. Specifically, we choose $K^\ell$ that can explain over $80\%$ of the total variance for each
stimulus and each layer, and therefore the value of $K^\ell$ varies with layer and input stimulus. 

\begin{figure}[h]
\centering
\includegraphics[bb=0 2 416 283,scale = 0.6]{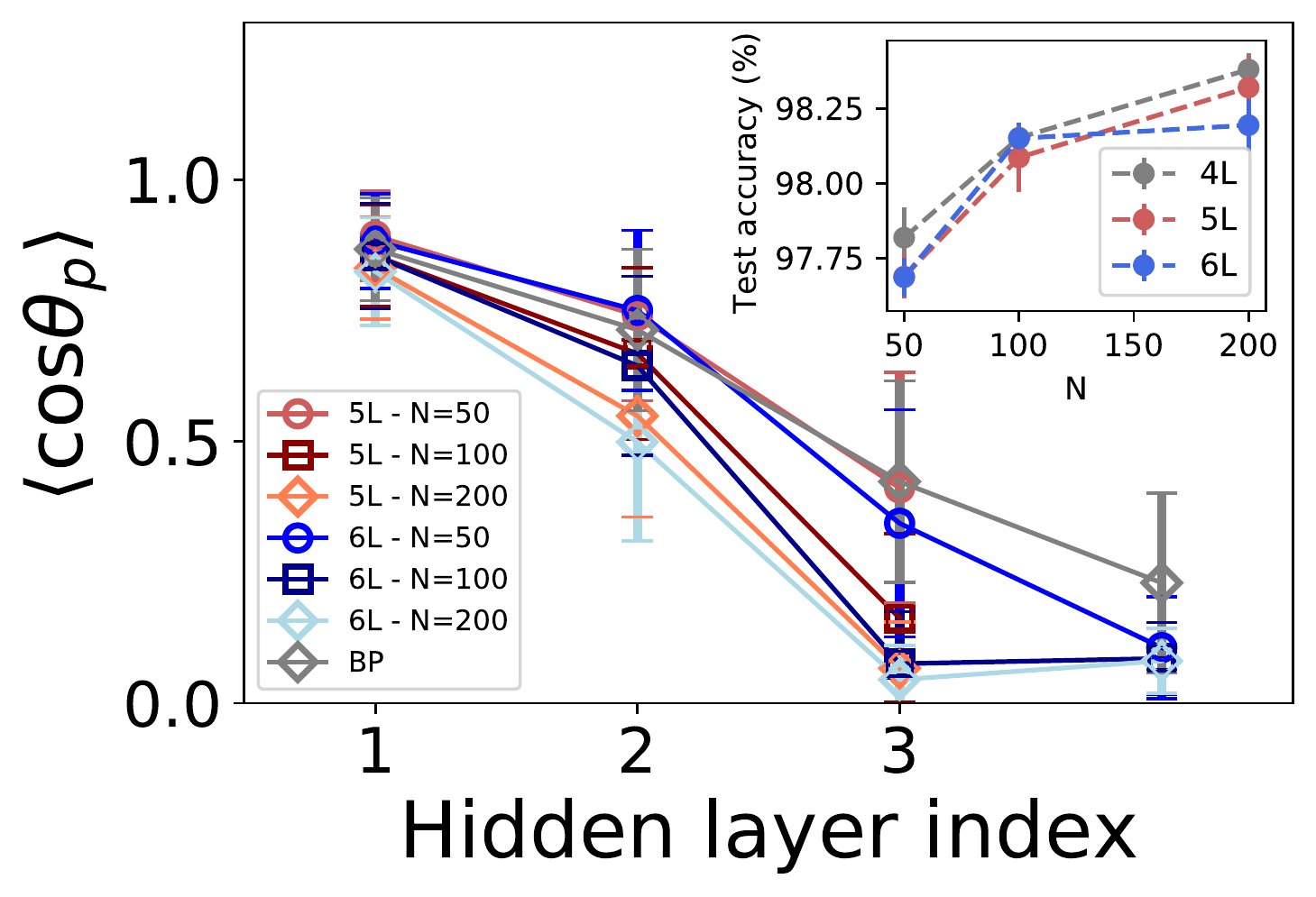}
\caption{The averaged subspace overlap versus layers. Different number of layers with different hidden-layer widths are considered. The results are averaged over five independent trainings. The inset
shows the corresponding test accuracy changing with the hidden-layer width.}
\label{figS10}
\end{figure}

In the main text, we observe a mild increase of the subspace overlap in deep layers. Here, as shown in Fig.~\ref{figS10}, we link this behavior to the saturation of the test performance with increasing number of layers and network width.
In addition, the task we consider is relatively simple, and thus three hidden layers (five layers in total) are sufficient to classify the digits with a high accuracy.
The subspace overlap under the MDL setting thus suggests a consistent way to determine the optimal number of layers and the network width in practical training.

\begin{figure}
	\centering
	\includegraphics[bb=4 2 564 414,width=1.0\textwidth]{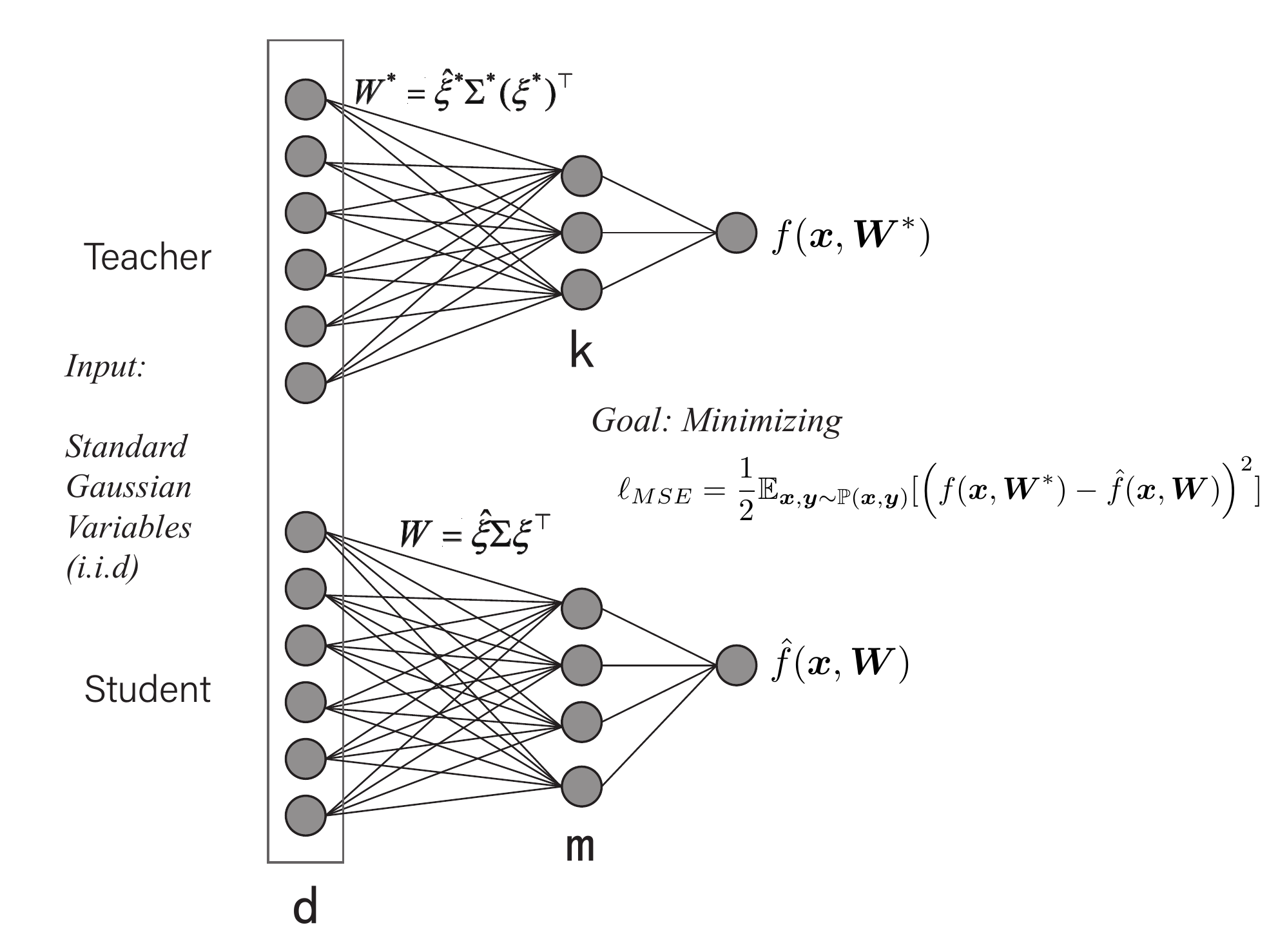}
	\caption{A simple illustration of the teacher-student setup
	with i.i.d. standard Gaussian input. The teacher network has $k$ 
	hidden nodes and $p^*$ modes, while the student network has $m$ hidden nodes and $p$ modes.
	The goal of the student network is to predict the labels generated by 
	the teacher network, minimizing the mean-squared error. The weights to the output layer are set to
	one in the linear readout for both teacher and student networks ($\mathbf{v}=\mathbf{I}_m, \mathbf{v}^*=\mathbf{I}_k$, and $\mathbf{I}_x$ is an all-one vector of length $x$).
	}\label{figS1}
\end{figure}

 \section{Mean-field predictions of on-line learning dynamics}
 In this section, we give a detailed derivation of the mean-field ordinary differential equations for 
the on-line dynamics. A sketch of the toy model setting is shown in Fig.~\ref{figS1}. The label for each sample $\mathbf{x}^{\nu}$ (i.i.d. standard
Gaussian variable) is generated by the teacher network,
\begin{equation}
y^{\nu}=f(\boldsymbol{\lambda}^{*\nu})=\frac{1}{k} \sum_{r=1}^{k} \sigma\left(\frac{[\boldsymbol{{\xi}}^*\mathbf{{\Sigma}^{*}}(\hat{\boldsymbol{\xi}}^{*})^{\t}]_{r} \mathbf{x}^{\nu}}{\sqrt{d}}\right)=\frac{1}{k} \sum_{r=1}^{k} \sigma\left(\lambda_{r}^{* \nu}\right), 
\end{equation}
where $[\mathbf{A}]_{r}$ 
denotes the $r$-th row of the matrix $\mathbf{A}$,
and $\lambda_r^{*\nu} = [\boldsymbol{{\xi}}^*\boldsymbol{{\Sigma}}^{*}(\hat{\boldsymbol\xi}^{*})^{\t}]_{r}\mathbf{x}^{\nu}/{\sqrt{d}}$ 
represents the $r$-th element of the teacher local field vector $\boldsymbol{\lambda}^{* \nu} \in \mathbb{R}^{k}$. 
To ensure the local field is independent of the input dimension,
we choose the initialization scheme for the teacher network
such that $[\boldsymbol{{\xi}}^*\boldsymbol{\Sigma}^{*}(\hat{\boldsymbol\xi}^{*})^{\t}]_{ij}\sim \mathcal{O}(1)$.
More precisely, we set the elements $\xi^*_{ik},\Sigma^*_{k},\hat{\xi}^*_{jk}$ to be independent standard
Gaussian variables, and then multiply the weight values by a factor of $\frac{1}{\sqrt{\ln d}}$ for logarithmic increasing number of modes.
This scaling ensures that the magnitude of the weight values is of the order one.
Different 
forms of transfer function $\sigma(\cdot)$ can be considered, but we choose the error function for the simplicity of the following theoretical analysis.
The prediction of the label by the student
network for a new sample $\mathbf{x}$ is given by
\begin{equation}
	\hat{f}(\mathbf{x}, \boldsymbol{\hat{\xi}},\mathbf{{\Sigma}},\boldsymbol{{\xi}})=\frac{1}{m} \sum_{r=1}^{m} \sigma\left(\frac{[\boldsymbol{\xi}\boldsymbol{{\Sigma}}(\hat{\boldsymbol\xi})^{\t}]_{r} \mathbf{x}}{\sqrt{d}}\right)=\frac{1}{m} \sum_{r=1}^{m} \sigma\left(\lambda_{r}\right),
\end{equation}
where $\lambda_{r}$ denotes the $r$-th 
component of the student local field
$\boldsymbol{\lambda} =\boldsymbol{\xi}\boldsymbol{\Sigma}\hat{\boldsymbol\xi}^{\t}\mathbf{x}$.
The student network has $m$ hidden nodes and $p$ patterns. For simplicity, we assume $m=k$, $p=p^*$, and only the pattern $\hat{\boldsymbol\xi}$ is learned.

Training the student network with the one-pass gradient descent (on-line learning) directly minimizes the following mean-squared error (MSE):
\begin{equation}
	\begin{aligned}
		&\ell_{MSE}(\boldsymbol{\lambda},\boldsymbol{\lambda}^*)=\frac{1}{2}\langle \left(\hat{f}(\boldsymbol{\lambda})-f\left(\boldsymbol{\lambda}^*\right)\right)^{2}\rangle,\\
	\end{aligned}	
\end{equation}
where $\langle \cdot \rangle$ indicates the average over $\{\mathbf{x,y}\}$ that can be replaced by the average over local fields.
For the Gaussian data $\mathbb{P}(\mathbf{x}) = \mathcal{N} (\mathbf{x}|0, \id)$,
the dynamics of $\ell_{MSE}$ can be completely determined by the following order parameters: $\mathbf{Q}^{\nu} \equiv \mathbb{E}_{\mathbf{x}, \mathbf{y} \sim \mathbb{P}(\mathbf{x}, \mathbf{y})}\left[\boldsymbol{\lambda}^{\nu} (\boldsymbol{\lambda}^{\nu})^{\t}\right]
=\frac{1}{d} \boldsymbol{\xi}^\nu\boldsymbol{\Sigma}^\nu (\boldsymbol{\hat{\xi}}^\nu)^{\t}\boldsymbol{\hat{\xi}}^\nu\boldsymbol{\Sigma}^\nu (\boldsymbol{\xi}^\nu)^{\t}$, 
$\mathbf{M}^{\nu} \equiv \mathbb{E}_{\mathbf{x}, \mathbf{y} \sim \mathbb{P}(\mathbf{x}, \mathbf{y})}\left[\boldsymbol{\lambda}^{\nu} (\boldsymbol{\lambda}^{* \nu})^{\t}\right]
=\frac{1}{d} \boldsymbol{\xi}^\nu\boldsymbol{\Sigma}^\nu (\boldsymbol{\hat{\xi}}^\nu)^{\t}\boldsymbol{\hat{\xi}}^{* \nu}\boldsymbol{\Sigma}^{* \nu} (\boldsymbol{\xi}^{* \nu})^{\t}$, 
and $\mathbf{P}^{\nu} \equiv \mathbb{E}_{\mathbf{x}, \mathbf{y} \sim \mathbb{P}(\mathbf{x}, \mathbf{y})}\left[\boldsymbol{\lambda}^{* \nu} (\boldsymbol{\lambda}^{* \nu})^{ \t}\right]
=\frac{1}{d} \boldsymbol{\xi}^{* \nu}\boldsymbol{\Sigma}^{* \nu} (\boldsymbol{\hat{\xi}}^{* \nu})^{\t}\boldsymbol{\hat{\xi}}^{* \nu}\boldsymbol{\Sigma}^{* \nu} (\boldsymbol{\xi}^{* \nu})^{\t}$. 
The corresponding matrix elements are denoted as $q_{j l}^{\nu} \equiv\left[\mathbf{Q}^{\nu}\right]_{j l}$, $m_{j r}^{\nu} \equiv\left[\mathbf{M}^{\nu}\right]_{j r}$ and $\rho_{r s} \equiv[\mathbf{P}]_{r s}$. Then we can define the local-field covariance
matrix $\mathbf{\Omega}^{\nu} \in \mathbb{R}^{(k+m) \times(k+m)}$ at time step $\nu$ as follows,
\begin{equation}
	\mathbf{\Omega}^{\nu} \equiv\left[\begin{array}{cc}
		\mathbf{Q}^{\nu} & \mathbf{M}^{\nu} \\
		(\mathbf{M}^{\nu})^{ \t} & \mathbf{P}
	\end{array}\right],
\end{equation}
where $\mathbf{P}$ is fixed by definition (parameters of the teacher network are quenched), the sample index $\nu$ is also the time step in the on-line learning setting,
and the evolution of other order parameters $\mathbf{Q}^{\nu}$ and $\mathbf{M}^{\nu}$ is driven by the gradient flow
of the mode parameters $\boldsymbol{\theta}^l = (\hat{\boldsymbol\xi}^l, \boldsymbol{\Sigma}^l, \boldsymbol{{\xi}}^{l+1})$. The loss is
completely determined by the evolving order parameters,
\begin{equation}
\begin{aligned}
	&\ell_{\mathrm{MSE}}(\boldsymbol\Omega)=\frac{1}{2}\mathbb{E}_{\boldsymbol{\lambda},\boldsymbol{\lambda}^{*} \sim \mathcal{N}\left(\boldsymbol{\lambda}, \boldsymbol{\lambda}^{*} \mid 0, \boldsymbol{\Omega}\right)}\left[\left(\hat{f}(\boldsymbol{\lambda})-f\left(\boldsymbol{\lambda}^{*}\right)\right)^{2}\right]\\
	&={\ell}_{\mathrm{t}}(\mathbf{P})+{\ell}_{\mathrm{s}}(\mathbf{Q})+\ell_{\mathrm{st}}(\mathbf{P}, \mathbf{Q}, \mathbf{M}),\
\end{aligned}	
\end{equation}
where
\begin{equation}
	\begin{aligned}
		&{\ell}_{\mathrm{t}}(\mathbf{P}) \equiv \mathbb{E}_{\boldsymbol{\lambda}^{*} \sim \mathcal{N}\left(\boldsymbol{\lambda}^{*} \mid 0, \mathbf{P}\right)}\left[f\left(\boldsymbol{\lambda}^{*}\right)^{2}\right]=\frac{1}{k^{2}} \sum_{r, s=1}^{k} \mathbb{E}_{\boldsymbol{\lambda}^{*} \sim \mathcal{N}\left(\boldsymbol{\lambda}^{*} \mid 0, \mathbf{P}\right)}\left[\sigma\left({\lambda}_{r}^{*}\right) \sigma\left({\lambda}_{s}^{*}\right)\right],\\
		&{\ell}_{\mathrm{s}}(\mathbf{Q}) \equiv \mathbb{E}_{\boldsymbol{\lambda} \sim \mathcal{N}(\boldsymbol{\lambda} \mid 0, \mathbf{Q})}\left[\hat{f}(\boldsymbol{\lambda})^{2}\right]=\frac{1}{m^{2}} \sum_{j, l=1}^{m} \mathbb{E}_{\boldsymbol{\lambda} \sim \mathcal{N}(\boldsymbol{\lambda} \mid 0, \mathbf{Q})}\left[\sigma\left(\lambda_{j}\right) \sigma\left(\lambda_{l}\right)\right] \text {, }\\
		&\ell_{\mathrm{st}}(\mathbf{P}, \mathbf{Q}, \mathbf{M}) \equiv \mathbb{E}_{\boldsymbol{\lambda}, \boldsymbol{\lambda}^{*} \sim \mathcal{N}\left(\boldsymbol{\lambda}, \boldsymbol{\lambda}^{*} \mid 0,\mathbf{ \Omega}\right)}\left[\hat{f}(\boldsymbol{\lambda}) f\left(\boldsymbol{\lambda}^{*}\right)\right]=-\frac{2}{m k} \sum_{j=1}^{m} \sum_{r=1}^{k} \mathbb{E}_{\boldsymbol{\lambda}, \boldsymbol{\lambda}^{*} \sim \mathcal{N}\left(\boldsymbol{\lambda}, \boldsymbol{\lambda}^{*} \mid 0, \mathbf{\Omega}\right)}\left[\sigma\left(\lambda_{j}\right) \sigma\left(\lambda_{r}^{*}\right)\right].
	\end{aligned}
\end{equation}
To proceed, we define the integral
$I_2 = \mathbb{E}_{\boldsymbol{\lambda}, \boldsymbol{\lambda}^{*} \sim \mathcal{N}\left(\boldsymbol{\lambda}, \boldsymbol{\lambda}^{*} \mid 0, \mathbf{\Omega}\right)}\left[\sigma\left(\boldsymbol{\lambda}^{\alpha}\right) \sigma\left(\boldsymbol{\lambda}^{\beta}\right)\right]$, which 
has an analytic form for $\sigma(x)=\operatorname{erf}(x / \sqrt{2})$ as follows,
\begin{equation}
	\mathbb{E}_{\boldsymbol{\lambda}, \boldsymbol{\lambda}^{*} \sim \mathcal{N}\left(\boldsymbol{\lambda}, \boldsymbol{\lambda}^{*} \mid 0, \mathbf{\Omega}\right)}\left[\sigma\left(\boldsymbol{\lambda}^{\alpha}\right) \sigma\left(\boldsymbol{\lambda}^{\beta}\right)\right]=\frac{2}{\pi} \arcsin \left(\frac{\Omega_{12}^{\alpha \beta}}{\sqrt{\left(1+\Omega_{11}^{\alpha \beta}\right)\left(1+\Omega_{22}^{\alpha \beta}\right)}}\right),
\end{equation}
where $\Omega_{ij}^{\alpha \beta}$ denotes the element of the overlap matrix for $\boldsymbol{\lambda}^{\alpha}$ and $\boldsymbol{\lambda}^{\beta}$, in which
$\alpha$ and $\beta$ indicate the attributes of the network---teacher or student. 
Therefore, the generalization error can be estimated as follows,
\begin{equation}
	\begin{aligned}
		\begin{aligned}
			\ell_{MSE}(\mathbf{\Omega})
			& = \frac{1}{k^{2}} \sum_{r, s=1}^{k} \frac{1}{\pi} \arcsin \left(\frac{\rho_{r s}}{\sqrt{\left(1+\rho_{r r}\right)\left(1+\rho_{s s}\right)}}\right)+\frac{1}{m^{2}} \sum_{j, l=1}^{m} \frac{1}{\pi} \arcsin \left(\frac{q_{j l}}{\sqrt{\left(1+q_{j j}\right)\left(1+q_{l l}\right)}}\right)\\&-\frac{2}{m k} \sum_{j=1}^{m} \sum_{r=1}^{k} \frac{1}{\pi} \arcsin \left(\frac{m_{j r}}{\sqrt{\left(1+q_{j j}\right)\left(1+\rho_{r r}\right)}}\right).
		\end{aligned}
	\end{aligned}
\end{equation}

We next consider the evolution of the order parameters, which involves only the update of $\boldsymbol{\hat{\xi}}$ in our toy model setting. 
Therefore, we derive the evolution of order parameters $q^{\nu}_{jl}$ and $m^{\nu}_{jr}$
based on the gradient of $\hat{\boldsymbol{\xi}}$: $\Delta\hat{\xi}_{j\alpha} =-\eta\frac{\hat{f}(\boldsymbol{\lambda}) - {f}(\boldsymbol{\lambda}^*)}{m\sqrt{d}}\sum_i \sigma^{\prime}(\lambda_i)\xi_{i\alpha}\Sigma_{\alpha}x_{j}^\nu$.
In the high-dimensional limit ($d\to\infty$), we use the self-averaging property of the order parameters considering the disorder average over the input data distribution~\cite{Biehl-1995,Saad-1995,Goldt-2019}.
Then we have the following expressions,
\begin{equation}
	\begin{aligned}
		q_{jl}^{\nu+1} - q_{jl}^{\nu} &= \frac{1}{d}\mathbb{E}\left[\sum_{n,\alpha,\beta}\xi_{j\alpha}\Sigma_\alpha (\hat{\xi}_{n\alpha}+\Delta \hat{\xi}_{n\alpha})(\hat{\xi}_{n\beta}+\Delta \hat{\xi}_{n\beta})\Sigma_\beta\xi_{l\beta}\right]- q_{jl}^{\nu},\\
		m_{jr}^{\nu+1} - m_{jr}^{\nu} & = \frac{1}{d}\mathbb{E}\left[\sum_{n,\alpha,\beta}\xi_{j\alpha}\Sigma_\alpha (\hat{\xi}_{n\alpha}+\Delta \hat{\xi}_{n\alpha})\hat{\xi}^*_{n\beta}\Sigma^*_\beta\xi^*_{r\beta}\right]- m_{jr}^{\nu},
	\end{aligned}
\end{equation}
where the expectation is carried out with respect to the data distribution.

Inserting the update equation of $\hat{\boldsymbol{\xi}}$ into the equation of the order parameter $q_{jl}$, we get
\begin{equation}\label{qdyn}
	\begin{aligned}
		q_{jl}^{\nu+1} - q_{jl}^{\nu} &= \frac{1}{d}\mathbb{E}\left[\sum_{n,\alpha,\beta}\xi_{j\alpha}\Sigma_\alpha (\hat{\xi}_{n\alpha}+\eta\frac{{f}(\boldsymbol{\lambda}^*)-\hat{f}(\boldsymbol{\lambda})}{m\sqrt{d}}\sum_i \sigma^{\prime}(\lambda_i)\xi_{i\alpha}\Sigma_{\alpha}x_{n}^\nu)(\hat{\xi}_{n\beta}+\eta\frac{{f}(\boldsymbol{\lambda}^*)-\hat{f}(\boldsymbol{\lambda})}{m\sqrt{d}}\sum_i \sigma^{\prime}(\lambda_i)\xi_{i\beta}\Sigma_{\beta}x_{n}^\nu)\Sigma_\beta\xi_{l\beta}\right]\\&- q_{jl}^{\nu},\\
		& = \frac{1}{d}\mathbb{E}\left[\sum_{n,\alpha,\beta}\xi_{j\alpha}\Sigma_\alpha \hat{\xi}_{n\alpha}\eta\frac{{f}(\boldsymbol{\lambda}^*)-\hat{f}(\boldsymbol{\lambda})}{m\sqrt{d}}\sum_i \sigma^{\prime}(\lambda_i)\xi_{i\beta}\Sigma_{\beta}x_{n}^\nu\Sigma_\beta\xi_{l\beta}\right]\\
		&+\frac{1}{d}\mathbb{E}\left[\sum_{n,\alpha,\beta}\xi_{j\alpha}\Sigma_\alpha \eta\frac{{f}(\boldsymbol{\lambda}^*)-\hat{f}(\boldsymbol{\lambda})}{m\sqrt{d}}\sum_i \sigma^{\prime}(\lambda_i)\xi_{i\alpha}\Sigma_{\alpha}x_{n}^\nu\hat{\xi}_{n\beta}\Sigma_\beta\xi_{l\beta}\right]\\
		&+\frac{1}{d}\mathbb{E}\left[\sum_{n,\alpha,\beta}\xi_{j\alpha}\Sigma_\alpha \eta\frac{{f}(\boldsymbol{\lambda}^*)-\hat{f}(\boldsymbol{\lambda})}{m\sqrt{d}}\sum_i \sigma^{\prime}(\lambda_i)\xi_{i\alpha}\Sigma_{\alpha}x_{n}^\nu(\eta\frac{{f}(\boldsymbol{\lambda}^*)-\hat{f}(\boldsymbol{\lambda})}{m\sqrt{d}}\sum_i \sigma^{\prime}(\lambda_i)\xi_{i\beta}\Sigma_{\beta}x_{n}^\nu)\Sigma_\beta\xi_{l\beta}\right],
	\end{aligned}
\end{equation}
where we have applied the definition of $q_{jl}^{\nu} = \frac{1}{d}\sum_{n,\alpha, \beta}\xi_{j\alpha}\Sigma_{\alpha}\hat{\xi}_{n\alpha}\hat{\xi}_{n\beta}\Sigma_{\beta}\xi_{l\beta}$ to derive the second equality.
Considering the definition of $\hat{f}(\boldsymbol{\lambda})$, ${f}(\boldsymbol{\lambda}^*)$,
$\boldsymbol{\lambda}$, and $\boldsymbol{\lambda}^*$, we recast Eq.~(\ref{qdyn}) as follows,
\begin{equation}
	\begin{aligned}
		q_{jl}^{\nu+1} - q_{jl}^{\nu} & = \frac{\eta}{dk m}\mathbb{E}\left[\sum_{\beta,r,i}\lambda_j \sigma\left(\lambda_{r}^{* \nu}\right) \sigma^{\prime}(\lambda_i)\xi_{i\beta}\Sigma_{\beta}^2\xi_{l\beta}\right]- \frac{\eta}{dm^2} \mathbb{E}\left[\sum_{\beta,\hat{r},i}\lambda_j \sigma\left(\lambda_{\hat{r}}^{ \nu}\right) \sigma^{\prime}(\lambda_i)\xi_{i\beta}\Sigma_{\beta}^2\xi_{l\beta}\right]\\
		&+\frac{\eta}{dk m}\mathbb{E}\left[\sum_{\alpha,r,i}\xi_{j\alpha}\Sigma_\alpha^2 \xi_{i\alpha}\lambda_l \sigma\left(\lambda_{r}^{* \nu}\right) \sigma^{\prime}(\lambda_i)\right]-\frac{\eta}{dm^2}\mathbb{E}\left[\sum_{\alpha,\hat{r},i}\xi_{j\alpha}\Sigma_\alpha^2 \xi_{i\alpha}\lambda_l \sigma\left(\lambda_{\hat{r}}^{ \nu}\right) \sigma^{\prime}(\lambda_i)\right]\\
		&+\frac{\eta^2}{dm^4}\mathbb{E}\left[\sum_{\alpha,\beta,i,a,r,\hat{r}}\xi_{j\alpha}\Sigma_\alpha^2 \xi_{i\alpha}\sigma^{\prime}(\lambda_i)\sigma^{\prime}(\lambda_a)\sigma\left(\lambda_{r}^{* \nu}\right)\sigma\left(\lambda_{\hat{r}}^{* \nu}\right)\xi_{{a\beta}}\Sigma_{\beta}^2\xi_{l\beta}
		 \right]\\
		&+\frac{\eta^2}{dm^2k^2}\mathbb{E}\left[\sum_{\alpha,\beta,i,a,r,\hat{r}}\xi_{j\alpha}\Sigma_\alpha^2 \xi_{i\alpha}\sigma^{\prime}(\lambda_i)\sigma^{\prime}(\lambda_a)\sigma\left(\lambda_{r}^{ \nu}\right)\sigma\left(\lambda_{\hat{r}}^{ \nu}\right)\xi_{{a\beta}}\Sigma_{\beta}^2\xi_{l\beta}
		\right]\\
		&-2\frac{\eta^2}{dm^3 k}\mathbb{E}\left[\sum_{\alpha,\beta,i,a,r,\hat{r}}\xi_{j\alpha}\Sigma_\alpha^2 \xi_{i\alpha}\sigma^{\prime}(\lambda_i)\sigma^{\prime}(\lambda_a)\sigma\left(\lambda_{r}^{* \nu}\right)\sigma\left(\lambda_{\hat{r}}^{ \nu}\right)\xi_{{a\beta}}\Sigma_{\beta}^2\xi_{l\beta},
		\right].
	\end{aligned}
\end{equation}
To proceed, we have to estimate the integral defined by
$I_3(\alpha,\beta,\eta) = \mathbb{E}_{\boldsymbol{\lambda}, \boldsymbol{\lambda}^{*} \sim \mathcal{N}\left(\boldsymbol{\lambda}, \boldsymbol{\lambda}^{*} \mid 0, \mathbf{\Omega}\right)}\left[\sigma^{\prime}\left(\boldsymbol{\lambda}^{\alpha}\right) \boldsymbol{\lambda}^{\beta} \sigma\left(\boldsymbol{\lambda}^{\eta}\right)\right]=\frac{2}{\pi} \frac{\Omega_{23}^{\alpha \beta \eta}\left(1+\Omega_{11}^{\alpha \beta \eta}\right)-\Omega_{12}^{\alpha \beta \eta} \Omega_{13}^{\alpha \beta \eta}}{\left(1+\Omega_{11}^{\alpha \beta \eta}\right) \sqrt{\left(1+\Omega_{11}^{\alpha \beta \eta}\right)\left(1+\Omega_{33}^{\alpha \beta \eta}\right)-\left(\Omega_{13}^{\alpha \beta \eta}\right)^{2}}}$ 
for our transfer function $\sigma(x)=\operatorname{erf}(x / \sqrt{2})$, where
$\Omega_{ij}^{\alpha \beta \eta}$ denotes the element of the field-covariance matrix
$\mathbf{\Omega}^{\alpha \beta \eta}$. We also have to estimate the second integral defined by
$I_4(i,j,k,l) = \mathbb{E}_{\boldsymbol{\lambda}, \boldsymbol{\lambda}^{*} \sim \mathcal{N}\left(\boldsymbol{\lambda}, \boldsymbol{\lambda}^{*} \mid 0, \mathbf{\Omega}\right)}\left[\sigma^{\prime}(\lambda_i)\sigma^{\prime}(\lambda_j)\sigma(\lambda_k)\sigma(\lambda_{l})\right]$, which has a closed form as
\begin{equation}
	I_4(i,j,k,l)=\frac{4}{\pi^{2}} \frac{1}{\sqrt{\bar{\Omega}_{0}^{ijkl}}} \arcsin \left(\frac{\bar{\Omega}_{1}^{ijkl}}{\sqrt{\bar{\Omega}_{2}^{ijkl} \bar{\Omega}_{3}^{ijkl}}}\right),
\end{equation}
where 
\begin{equation}
	\begin{array}{c}
		\bar{\Omega}_{0}^{i j k l} \equiv\left(1+\Omega_{11}^{i j k l}\right)\left(1+\Omega_{22}^{i j k l}\right)-\left(\Omega_{12}^{i j k l}\right)^{2}, \\
		\bar{\Omega}_{1}^{i j k l} \equiv \bar{\Omega}_{0}^{i j k l} \Omega_{34}^{i j k l}-\Omega_{23}^{i j k l} \Omega_{24}^{i j k l}\left(1+\Omega_{11}^{i j k l}\right)-\Omega_{13}^{i j k l} \Omega_{14}^{i j k l}\left(1+\Omega_{22}^{i j k l}\right) \\
		+\Omega_{12}^{i j k l} \Omega_{13}^{i j k l} \Omega_{24}^{i j k l}+\Omega_{12}^{i j k l} \Omega_{14}^{i j k l} \Omega_{23}^{i j k l}, \\
		\bar{\Omega}_{2}^{i j k l} \equiv \bar{\Omega}_{0}^{i j k l}\left(1+\Omega_{44}^{i j k l}\right)-\left(\Omega_{24}^{i j k l}\right)^{2}\left(1+\Omega_{11}^{i j k l}\right)-\left(\Omega_{13}^{i j k l}\right)^{2}\left(1+\Omega_{22}^{i j k l}\right)+2 \Omega_{12}^{i j k l} \Omega_{13}^{i j k l} \Omega_{23}^{i j k l}, \\
		\bar{\Omega}_{3}^{i j k l} \equiv \bar{\Omega}_{0}^{i j k l}\left(1+\Omega_{44}^{i j k l}\right)-\left(\Omega_{24}^{i j k l}\right)^{2}\left(1+\Omega_{11}^{i j k l}\right)-\left(\Omega_{14}^{i j k l}\right)^{2}\left(1+\Omega_{22}^{i j k l}\right)+2 \Omega_{12}^{i j k l} \Omega_{14}^{i j k l} \Omega_{24}^{i j k l}.
	\end{array}
\end{equation}

In an analogous way, we can derive the mean-field evolution of $m_{jr}^{\nu}$ as follows,
\begin{equation}
	\begin{aligned}
			m_{jr}^{\nu+1} - m_{jr}^{\nu} & = \frac{1}{d}\mathbb{E}\left[\sum_{n,\alpha,\beta}\xi_{j\alpha}\Sigma_\alpha (\hat{\xi}_{n\alpha}+\Delta \hat{\xi}_{n\alpha})\hat{\xi}^*_{n\beta}\Sigma^*_\beta\xi^*_{r\beta}\right]- m_{jr}^{\nu}\\
			& = \frac{\eta}{k md}\mathbb{E}\left[\sum_{\alpha,i,a}\xi_{j\alpha}\Sigma_\alpha^2 \xi_{i\alpha}{\sigma(\lambda_a^*)} \sigma^{\prime}(\lambda_i)\lambda_r^*\right] - \frac{\eta}{m^2d}\mathbb{E}\left[\sum_{\alpha,i,a}\xi_{j\alpha}\Sigma_\alpha^2 \xi_{i\alpha}{\sigma(\lambda_a)} \sigma^{\prime}(\lambda_i)\lambda_r^*\right],
	\end{aligned}
\end{equation}
where the definition of $m_{jr}$ has bee used.
If we define $\tau\equiv \nu/d$, and take the thermodynamic limit of $d\to \infty$, the time step becomes continuous, and we can thus write down the following ODEs,
\begin{equation}
	\begin{aligned}
		\frac{\mathrm{d}q_{jl}}{\mathrm{d}\tau}& = \frac{\eta}{km}\left[\sum_{\beta,r^*,i}I_3(i,j,r^*)\xi_{i\beta}\Sigma_{\beta}^2\xi_{l\beta}\right]- \frac{\eta}{m^2} \left[\sum_{\beta,\hat{r},i}I_3(i,j,\hat{r})\xi_{i\beta}\Sigma_{\beta}^2\xi_{l\beta}\right]\\
		&+\frac{\eta}{km}\left[\sum_{\alpha,r^*,i}\xi_{j\alpha}\Sigma_\alpha^2 \xi_{i\alpha}I_3(i,l,r^*)\right]-\frac{\eta}{m^2}\mathbb{E}\left[\sum_{\alpha,\hat{r},i}\xi_{j\alpha}\Sigma_\alpha^2 \xi_{i\alpha}I_3(i,l,\hat{r})\right]\\
		&+\frac{\eta^2}{m^4}\left[\sum_{\alpha,\beta,i,a,r^*,\hat{r}^*}\xi_{j\alpha}\Sigma_\alpha^2 \xi_{i\alpha}I_4(i,a,r^*,\hat{r}^*)\xi_{a\beta}\Sigma_{\beta}^2\xi_{l\beta}
		\right]\\
		&+\frac{\eta^2}{m^2k^2}\left[\sum_{\alpha,\beta,i,a,r,\hat{r}}\xi_{j\alpha}\Sigma_\alpha^2 \xi_{i\alpha}I_4(i,a,r,\hat{r})\xi_{a\beta}\Sigma_{\beta}^2\xi_{l\beta}
		\right]\\
		&-2\frac{\eta^2}{m^3k}\left[\sum_{\alpha,\beta,i,a,r^*,\hat{r}}\xi_{j\alpha}\Sigma_\alpha^2 \xi_{i\alpha}I_4(i,a,r^*,\hat{r})\xi_{a\beta}\Sigma_{\beta}^2\xi_{l\beta}
		\right],\\
		\frac{\mathrm{d}m_{jr^*}^\nu}{\mathrm{d}\tau}& = \frac{\eta}{km}\left[\sum_{\alpha,i,a}\xi_{j\alpha}\Sigma_\alpha^2 \xi_{i\alpha}I_3(i,r^*,a^*)\right] - \frac{\eta}{m^2}\left[\sum_{\alpha,i,a}\xi_{j\alpha}\Sigma_\alpha^2 \xi_{i\alpha}I_3(i,r^*,a)\right],
	\end{aligned}
\end{equation}
where the index in $I_3$ or $I_4$ with the symbol $*$ labels the teacher's local-field.




\end{document}